\documentclass[10pt,letterpaper]{article}
\usepackage[top=0.85in,left=2.75in,footskip=0.75in]{geometry}

\usepackage{amsmath,amssymb}

\usepackage{changepage}

\usepackage[utf8x]{inputenc}

\usepackage{textcomp,marvosym}

\usepackage{cite}

\usepackage{nameref,hyperref}

\usepackage[right]{lineno}

\usepackage{microtype}
\DisableLigatures[f]{encoding = *, family = * }

\usepackage[table]{xcolor}

\usepackage{array}

\newcolumntype{+}{!{\vrule width 2pt}}

\newlength\savedwidth

\raggedright
\usepackage[aboveskip=1pt,labelfont=bf,labelsep=period,justification=raggedright,singlelinecheck=off]{caption}
\renewcommand{\figurename}{Fig}

\makeatletter
\renewcommand{\@biblabel}[1]{\quad#1.}
\makeatother

\date{}

\usepackage{lastpage,fancyhdr,graphicx}
\usepackage{epstopdf}

\newcommand{\beginsupplement}{%
        \setcounter{table}{0}
        \renewcommand{\thetable}{S\arabic{table}}%
        \setcounter{figure}{0}

        \renewcommand{\figurename}{\hspace{-3pt}}
        \renewcommand{\thefigure}{S\arabic{figure} Fig}   
        
        \renewcommand{\tablename}{\hspace{-3pt}}
        \renewcommand{\thetable}{S\arabic{table} Table}

     }

\begin{document}
\vspace*{0.2in}

\begin{flushleft}
{\Large
\textbf\newline{Diffusion-based neuromodulation can eliminate catastrophic forgetting in simple neural networks} 
}
\newline
\\
Roby Velez\textsuperscript{1},
Jeff Clune\textsuperscript{1,2*}
\\
\bigskip
\textbf{1} Computer Science Department, University of Wyoming, Laramie,
Wyoming, United States of America
\\
\textbf{2} Uber AI Labs, San Francisco,
California, United States of America
\\
\bigskip

* jeffclune@uwyo.edu
 
\end{flushleft}

\section*{Abstract}

A long-term goal of AI is to produce agents that can learn a diversity of skills throughout their lifetimes and continuously improve those skills via experience. A longstanding obstacle towards that goal is catastrophic forgetting, which is when learning new information erases previously learned information. Catastrophic forgetting occurs in artificial neural networks (ANNs), which have fueled most recent advances in AI. A recent paper proposed that catastrophic forgetting in ANNs can be reduced by promoting modularity, which can limit forgetting by isolating task information to specific clusters of nodes and connections (functional modules). While the prior work did show that modular ANNs suffered less from catastrophic forgetting, it was not able to produce ANNs that possessed task-specific functional modules, thereby leaving the main theory regarding modularity and forgetting untested. We introduce diffusion-based neuromodulation, which simulates the release of diffusing, neuromodulatory chemicals within an ANN that can modulate (i.e. up or down regulate) learning in a spatial region. On the simple diagnostic problem from the prior work, diffusion-based neuromodulation 1) induces task-specific learning in groups of nodes and connections (task-specific localized learning), which 2) produces functional modules for each subtask, and 3) yields higher performance by eliminating catastrophic forgetting. Overall, our results suggest that diffusion-based neuromodulation promotes task-specific localized learning and functional modularity, which can help solve the challenging, but important problem of catastrophic forgetting.


\section*{Introduction}

Learning is a powerful, complex ability possessed by natural organisms, and one that artificial intelligence researchers have sought to incorporate into artificial systems. Advances in learning systems such as deep neural networks (DNNs) have led to major innovations through state-of-the-art performances in vision recognition~\cite{krizhevsky2012imagenet}, video game playing~\cite{mnih2015human}, robot control~\cite{levine2016end} and many other domains~\cite{goodfellow2016deep}. While DNNs and other learning systems have become quite powerful in recent years they still lack a crucial aspect of natural learning: the ability to continuously learn new skills over a lifetime. Artificial learning systems are unable to continuously learn new information due to a phenomenon called \emph{catastrophic forgetting}. 

Catastrophic forgetting is when the learning of new information causes old information to be rapidly lost~\cite{french1992semi,mermillod2013stability}. It is particularly extreme in artificial neural networks (ANNs)~\cite{french1999catastrophic,mermillod2013stability}. ANNs are graph-based structures that are simplified, abstract computational models of real brains in which the nodes and connections of the graph correspond to neurons and synapses~\cite{Haykin1998,floreano2008bio}. Like real brains, information in ANNs is encoded in connection weights and patterns and learning involves the changing of those weights~\cite{haykin2009neural,widrow199030,rumelhart1988learning,soltoggio2008neural}. One reason ANNs are prone to catastrophic forgetting is because information for tasks tends to be spread across many nodes and connections, rather than isolated to specific groups of nodes and connections~\cite{french1999catastrophic}. In such a situation, any change to a group of nodes and connections to learn new information would cause forgetting because those nodes and connections most likely encoded for something else~\cite{gutstein2015reduction,christian1990cascade}. One possible solution is to encourage the isolation of information to specific groups of nodes and connections. This isolation should help disentangle the parts of the ANN that encode for different aspects of problems~\cite{french1992semi}.

Ellefsen et al.~\cite{ellefsen2015neural} have proposed that modularity could facilitate the isolation of information to specific groups of nodes and connections. Modularity within ANNs, and networks in general, is characterized by clusters of highly interconnected nodes (i.e. modules) that are sparsely connected to other clusters~\cite{wagner2007road,guimera2005functional,lipson2007principles}. Previous research showed that modular ANNs could be produced via a method known as the connection cost technique (CCT)~\cite{clune2013originModularity,huizinga2014evolving}. With the CCT, ANNs are evolved with an evolutionary algorithm (EA) that includes an evolutionary cost for each connection~\cite{clune2013originModularity}. EAs are search algorithms based on Darwinian evolution, and can search through various ANN configurations for the right weights that allow an ANN to solve a problem~\cite{floreano2008bio,yao1999evolving}. Modularity could facilitate learning to be turned on within a module without interfering with information in the rest of the ANN, and thus could reduce or eliminate catastrophic forgetting. This learning within a module should isolate information and produce \emph{functional modules} that encode for specific information, such as a subproblem or task in a multitask problem. While Ellefsen et al.~\cite{ellefsen2015neural} found that modular ANNs, produced via the CCT, suffered less from catastrophic forgetting, their ANNs did not possess functional modules for the different skills tested; thereby leaving the main tenet of their hypothesis untested. In this paper, we introduce a method based on diffusion that can produce the isolation of information in functional modules by inducing learning that corresponds to a specific subtask in a group of nodes and connections (i.e. task-specific localized learning).

Neurons employ a wide array of communication mechanisms. Traditionally, neuronal communication is viewed as a private channel or wire of communication between two neurons facilitated by a synapse or gap junction~\cite{levitan2015neuron}, and is sometimes referred to as \emph{wire transmission}. Neurons have also been shown to engage in \emph{volume transmission} where they release signaling chemicals, such as neurotransmitters, that can diffuse and transmit information to neurons within a volume of brain tissue~\cite{fuxe2007golgi,agnati2006volume}. Diffusing neurotransmitters can only influence the neurons in their general vicinity due to obstructions and recycling factors in the extracellular space (ECS) between neurons~\cite{nicholson1998extracellular}, and many play a role in synaptic plasticity and learning~\cite{bailey2000heterosynaptic,marder2002cellular,jay2003dopamine}. Because of these two properties, it is possible that volume transmission could be producing some \emph{localized learning}, where groups of neurons and synapses within a volume of brain tissue all undergo learning at the same time. It is difficult to assess whether this localized learning is task-specific because much of the brain's mechanisms and processes are still unknown. It has been suggested by many researchers, though mostly in passing, that the synchronized and coordinated learning in groups of neurons could play a role in the creation or maintenance of functional units or modules~\cite{engert1997synapse,diamond2002broad,schoppa2001glomerulus,fuxe2012extrasynaptic,agnati2006volume}.

In this paper, we abstract the idea of volume transmission via diffusing chemical signals in real brains to produce a new learning algorithm for ANNs called diffusion-based neuromodulation. In this implementation of diffusion-based neuromodulation, we place point sources at specific locations within an ANN that emit diffusing learning signals that correspond to the positive and negative feedback for the tasks being learned. We test whether these diffusing learning signals can 1) induce task-specific localized learning in order to 2) isolate information for the different tasks into functional modules and 3) reduce catastrophic forgetting. 
\subsection*{Background}

\subsubsection*{Modularity}

Modularity is a important feature in both man-made~\cite{baldwin2000design} and natural systems~\cite{alon2006introduction,carroll2001chance,Hintze2008,wagner2007road}. One of the benefits of modularity is that it allows the components of a system to be easily reconfigured or replaced~\cite{baldwin2000design,Kashtan2005,carroll2001chance,klingenberg2005developmental}. In the context of ANNs, there is structural modularity and functional modularity. Structural modularity quantifies the connectivity pattern of nodes and connections, and is the most studied. Two methods to promote structural modularity during the evolution of an ANN include the CCT mentioned above, and constantly switching between different test problems that have the same subgoals~\cite{Kashtan2005}. Structural modularity in these works was quantified with the Q-Score metric~\cite{Leicht2008} which quantifies the connectivity patterns of nodes and connections, and is the current state-of-the-art in module detection. Functional modularity involves modules that encode for some specific information, such as a subproblem or one of the tasks in a multitask problem~\cite{lipson2007principles}. Identification of functional modules is challenging because it requires understanding how information is encoded in the nodes and connections of an ANN.

A recent paper presented a technique to identify functional modules called subsets regression on network connectivity (SRC)~\cite{velez2016identifying}. It identifies the nodes and connections that encode for subproblems of an overall task. The result is functional modules, and a core functional network, which is a subnetwork of the original ANN that has at least the same fitness. When the Q-Score metric is applied to a CFN it produces a \emph{functional modularity} score, i.e. a structural modularity score based only on the functional nodes and connections. Functional modularity and the ability to identify functional modules are crucial to the study of catastrophic forgetting in this paper because they allow us to understand how information for the different tasks is encoded in the ANNs.

\subsubsection*{Learning and Forgetting} 
\label{sec:learningForgeting}

Due to the complexity of even small ANNs, these structures can not be fully designed by hand and researchers must rely on automated methods to set their weights. The two most prominent approaches are EAs and learning algorithms. While quite powerful, the ANNs produced by EAs are generally static, and cannot further learn or incorporate information during their lifetime. In contrast, learning algorithms such as Hebbian learning~\cite{hebb2005organization}, neuromodulation~\cite{soltoggio2007evolving}, and backpropagation~\cite{haykin2009neural,widrow199030,rumelhart1988learning} enable ANNs to continuously learn during their lifetime. Many researchers combine both methods and evolve the starting weights for an ANN, and then incorporate learning to further refine the weights of the network~\cite{floreano2008bio,yao1999evolving}. The ANNs in this work, and in Ellefsen et al.~\cite{ellefsen2015neural}, implement this latter approach of combining evolution and learning.

In Hebbian learning the strength of a connection between nodes increases or decreases depending on whether the firing of those nodes is correlated or non-correlated~\cite{hebb2005organization}. Hebbian learning also occurs in neuromodulation, but in neuromodulation there is a mechanism that can modulate (i.e. raise, lower, or invert) the rate of Hebbian learning. In neuromodulation ANNs, there are two types of nodes: regular nodes and modulatory nodes. Through a direct connection to a regular node, a modulatory node can modulate the rate of Hebbian learning in the connections feeding into that regular node~\cite{soltoggio2007evolving}. Put another way, in neuromodulation, learning can be context specific because there is a mechanism to turn Hebbian learning on or off in target connections given specific situations or data. Hebbian learning and neuromodulation are modeled after homosynaptic and heterosynaptic plasticity rules found within real brains~\cite{bailey2000heterosynaptic}. Neuromodulation has been successful at training simulated bees in foraging tasks where the bees had to learn which flowers produced the highest reward~\cite{soltoggio2007evolving}. Neuromodulation has also been successful, more so than regular Hebbian learning, at creating robots that can navigate a maze filled with moving rewards~\cite{soltoggio2008evolutionary}. Neuromodulation was the learning algorithm in Ellefsen et al.~\cite{ellefsen2015neural}, and is the basis for diffusion-based neuromodulation.

Another ANN learning algorithm is backpropagation~\cite{rumelhart1988learning}. It differs from Hebbian learning and neuromodulation in that it requires knowing the correct ANN output for all inputs in order to calculate detailed error signals. Backpropagation then sends those error signals back through the ANN and applies weight changes to connections based on how much influence they had over those error signals. Backpropagation has been very successful at training DNNs and has fueled many of the major advances in AI in recent years~\cite{krizhevsky2012imagenet,mnih2015human,levine2016end,goodfellow2016deep}. DNNs can also suffer from catastrophic forgetting~\cite{goodfellow2013empirical}, although there has been some recent progress in this area~\cite{Kirkpatrick14032017}. If we can solve catastrophic forgetting on small diagnostic problems we could potentially scale those solutions up to DNNs and increase their capabilities.

In addition to Ellefsen et al.~\cite{ellefsen2015neural}, another method that can reduce catastrophic forgetting by isolating information to specific nodes and connections is node sharpening. During learning, node sharpening influences the weight changes for connections feeding into the most and least active nodes, making those nodes more and less active, respectively~\cite{french1992semi}. The end result is that only a few nodes and connections, not the entire ANN, encode for a specific task or piece of information. Researchers have also evolved ANNs that have the ability to write data to memory on disk and read it back later through evolvable neural turing machines (ENTMs)~\cite{luders2017continual}. ENTMs were applied to the foraging task, which is the experimental domain is this paper and Ellefsen et al.~\cite{ellefsen2015neural}, and produced a few individuals that completely avoided catastrophic forgetting, but were not able to reliably produce perfect solutions across all runs. Other strategies to combat catastrophic forgetting include rehearsing previously learned skills~\cite{ratcliff1990connectionist,robins1995catastrophic}, emulating dual memory models~\cite{french1992semi,hinton1987using}, or developing routines that determine which weights should become static and retain older tasks and which should stay plastic to learn a new task~\cite{Kirkpatrick14032017}.

\subsubsection*{Diffusion} 

A growing body of work is beginning to illuminate the prevalence of volume transmission in real brains, and show that neurons can engage in a mix of both wire transmission and volume transmission~\cite{fuxe2007golgi,agnati2010understanding}. One example of volume transmission is the \emph{spillover} of neurotransmitters like glutamate or gamma-Aminobutyric acid (GABA). Neurotransmitters can have many different functions in the brain, but glutamate and GABA are generally classified as messenger chemicals that can excite or inhibit a neuron~\cite{levitan2015neuron}. When transmitting a signal, a pre-synaptic neuron releases neurotransmitters from the vesicles at the end of its synapses that diffuse across the synaptic cleft to excite or inhibit the receiving, post-synaptic neuron~\cite{levitan2015neuron}.  The neurotransmitter usually stays within the synaptic cleft, but sometimes it can spillover into the extracellular space (ECS) and affect neurons in the surrounding area~\cite{diamond2002broad,isaacson2000synaptic,huang1998synaptic,sem2005diffusional}. Neurotransmitter spillover has been observed in different areas of the brain such as the hippocampus~\cite{kullmann1998extrasynaptic,engert1997synapse,scanziani2000gaba}, cerebellum~\cite{rossi1998spillover}, and olfactory bulb~\cite{schoppa2001glomerulus,isaacson1999glutamate}. Aside from spillover, neurons can also directly inject neurotransmitters, such as the neuromodulators dopamine~\cite{rice2011dopamine,descarries1996dual} and serotonin~\cite{de2005synaptic,descarries1975serotonin}, into the ECS, without any consideration towards targeting a particular neuron~\cite{trueta2012extrasynaptic,taber2014volume,fuxe2012extrasynaptic}. Generally, neuromodulators are classified as neurotransmitters that can modulate synaptic strength, and are central in models of heterosynaptic plasticity and learning within the brain~\cite{bailey2000heterosynaptic,marder2002cellular,jay2003dopamine}. The strongest evidence for this deliberate broadcasting of neurotransmitter into the ECS is the fact that in certain regions of the brain there are far more neurotransmitter receptors than transmitters~\cite{trueta2012extrasynaptic,taber2014volume,fuxe2012extrasynaptic}. Lastly, another mechanism for volume transmissions comes from gaseous neurotransmitters such as nitric oxide $(NO)$, carbon monoxide $(CO)$, and hydrogen sulfide $(H_{2}S)$~\cite{wang2002two}. NO is the most studied of these gaseous neurotransmitters and has been linked to synaptic plasticity and learning~\cite{garthwaite2008concepts,susswein2004nitric}. NO is a highly diffusible, molecular gas that can move easily through cell membranes, and simply starts to diffuse as soon as it is synthesized within a neuron~\cite{garthwaite2008concepts,esplugues2002no,dawson1994gases,gally1990no}. Due to its highly diffusible properties and effect on synaptic plasticity, NO has been abstracted to an ANN framework called GasNets that has been shown to be comparable to other ANN frameworks in regards to visual navigation~\cite{husbands1998better} and bipedal locomotion tasks~\cite{mchale2004gasnets}. Lastly, even with factors that limit a diffusing chemical signal such as obstructions and uptake in the ECS~\cite{nicholson1998extracellular} or general dilution~\cite{rice2008dopamine}, simulations of the diffusion of dopamine~\cite{rice2008dopamine}, glutamate~\cite{rusakov1998extrasynaptic}, and NO~\cite{wood1994models} indicate that these chemicals can diffuse far enough to affect large populations of neurons. 

\subsection*{Experimental Setup}

\begin{figure}[h!]
	\centering
	\vspace{-215pt}
         \includegraphics[width=\textwidth]{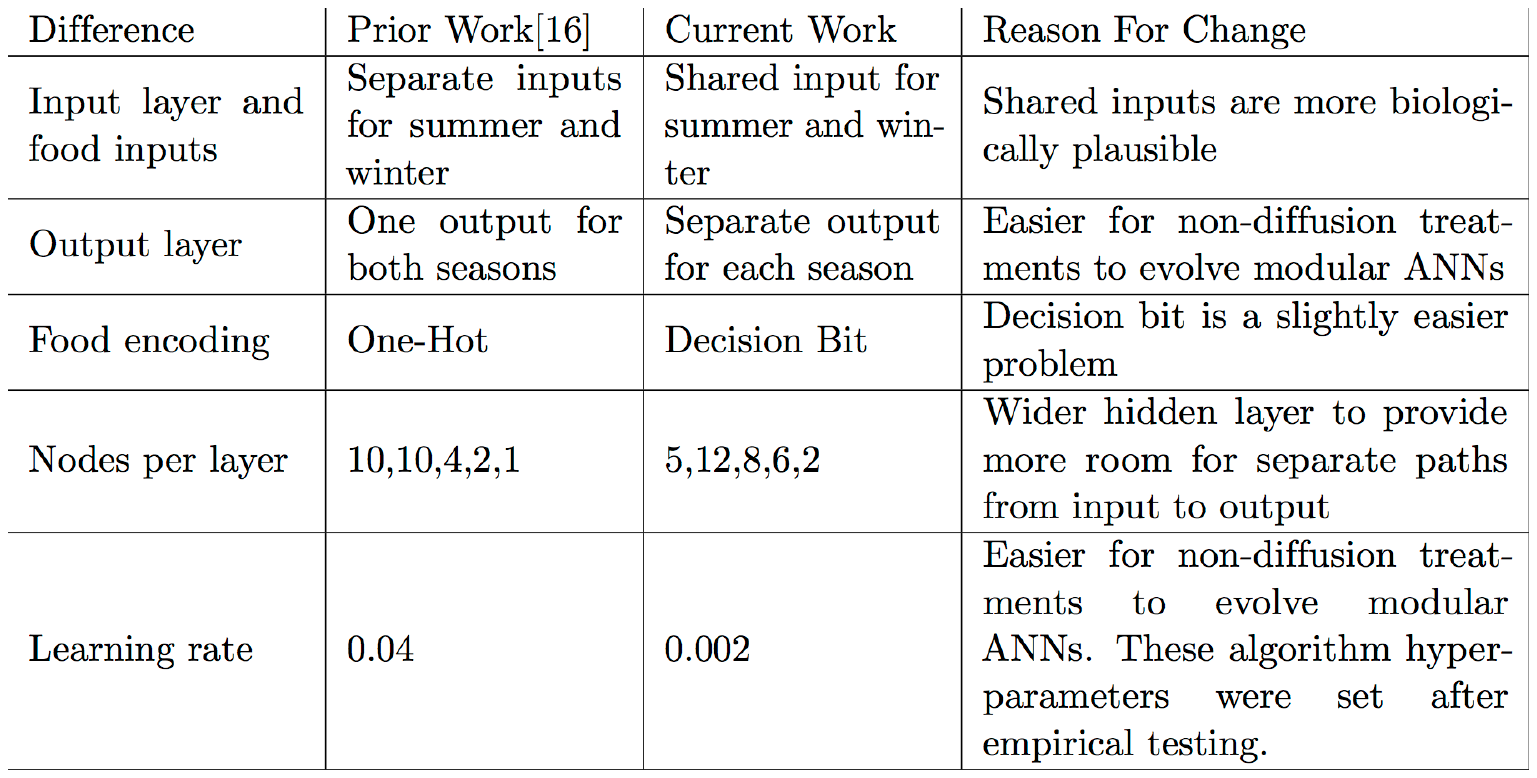}
	\vspace{-175pt}
        \caption{      
{\bf Differences between this work and Ellefsen et al.~\cite{ellefsen2015neural}.} Differences prevent direct comparison between the non-diffusion treatments in this work and the networks in Ellefsen et al.~\cite{ellefsen2015neural}. Purpose of many of the changes were to make it easier for modular solutions to appear in order to investigate whether they aid with catastrophic forgetting.}
\label{S1_Table}
\end{figure}

This section briefly describes the experimental setup in this paper designed to test catastrophic forgetting. A more detailed description of the implementation is in \nameref{sec:Methods}. With a few exceptions (\ref{S1_Table}) the experimental setup is the same as Ellefsen et al.~\cite{ellefsen2015neural}. Because the network topology, food encoding, and some of the learning parameters are different from Ellefsen et al.~\cite{ellefsen2015neural}, the networks from this paper can not be directly compared to their work. 

\subsubsection* {Foraging Task}
\label{sec:forageDescription}

We conduct experiments in a variant of the foraging task, introduced by Ellefsen et al.~\cite{ellefsen2015neural}, where an artificial agent is presented food items during a series of days. Each day the agent is presented with all possible food items and its task is to learn which food items are nutritious and should be eaten, and which are poisonous and should not be eaten. After five days the agent transitions to a new season where the food items are the same, but their association (nutritious or poisonous) is reassigned randomly. The seasons the agent experiences are summer and winter, and together they make up a year. The agent's lifetime is three years in total and the food associations for each particular season stay constant over that lifetime. Within each season, half of the food items are nutritious and half are poisonous. The summer and winter food associations, along with the order in which they are presented in a lifetime, are called an environment. To achieve maximum fitness, an agent must eat all the nutritious items and not eat the poisonous items (Eq~\ref{eq:fitnessFunction}). 

\begin{equation}
\label{eq:fitnessFunction}
fitness=0.5+\frac{nutritiousFoodEaten-poisonousFoodEaten}{totalFood}
\end{equation}

A successful agent is one that learns the correct food associations in the first season (i.e. summer) and then, when learning the correct associations in the second season (i.e. winter), does not forget what it learned in the prior season. For the remaining two years of the agent's lifetime, it can thus recall the associations it already knows to make the correct decisions. On the other hand, if the learning of food associations in one season causes the loss of associations for the other season, then, as the seasons cycle, the agent will have to continuously relearn associations again and again, which results in mistakes that lower fitness.

\subsubsection *{Network Setup and Encodings}

\begin{figure}[!h]
	\centering
         \includegraphics[width=\textwidth]{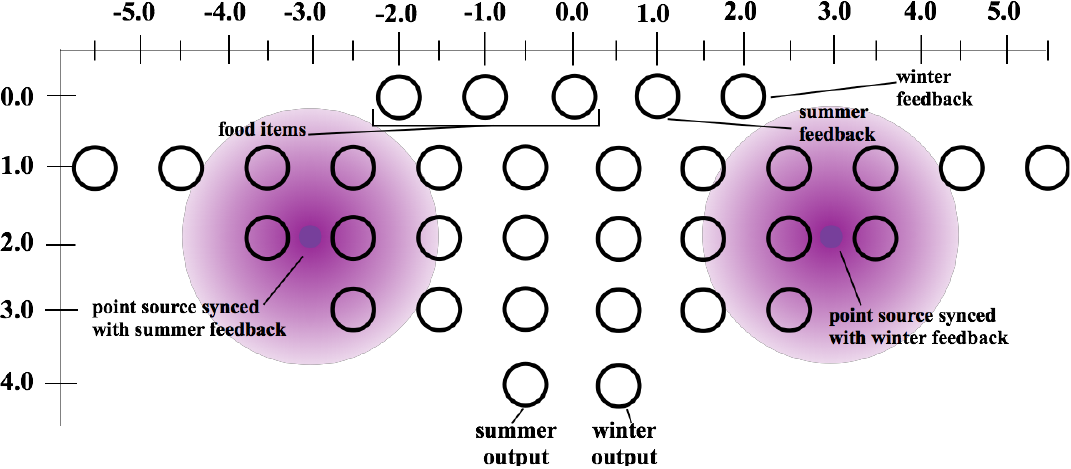}
        \caption{
        {\bf Network Topology} 
Individuals in the foraging task are represented as ANNs where each node possesses an $(x,y)$ position. The first three inputs correspond to food items while the last two inputs are fed positive $(1)$ and negative $(-1)$ feedback signals for the summer and winter season respectively.  An output greater than 0 results in the agent eating the food item presented. Two point sources, one for each season, exist at the locations $(-3,2)$, and $(3,2)$. Their activation is synchronized to the positive and negative feedback of the summer and winter season. They affect all nodes within a radius of 1.5 and the modulatory signal of the point sources increases as a Gaussian as you get closer (Eq~\ref{eq:gaussEquation}).
}
\label{S1_Fig}

\vspace{-50pt}

\end{figure}

The artificial agents are represented by feed-forward, five-layer networks where each node has an (x,y) position  (\ref{S1_Fig}). The number of nodes in each layer from input to output are 5, 12, 8, 6, and 2 respectively. Starting from left to right, the first three nodes in the input layer are fed food items (described below) for both seasons, and are referred to as a \emph{shared input}. The last two nodes are referred to as \emph{seasonal feedback} because they are fed feedback signals, and are season specific. The feedback is 0 if the previously presented food item was not eaten, and is 1 or -1 if the previously presented food item was eaten and it was nutritious or poisonous. The summer and winter feedback nodes are fed feedback during the summer and winter season respectively, and are inactive (i.e. fed 0) during the other season. Lastly, the two outputs are also season specific and determine whether the agent eats $(output>0)$ or does not eat the food item presented. In summer only the leftmost output is considered and in winter only the rightmost output is considered. 

The food items presented to the ANNs, and fed into the first 3 nodes of the input layer, are encoded as a 3-bit vector of $1$'s and $-1$'s. The food associations, whether something is nutritious or poisonous, for each season are randomly assigned when creating an environment. For each season, a bit in the food encoding is chosen at random to be the \emph{decision bit}. A coin flip is then done to determine whether encodings with a -1 or 1 in the decision bit signify a nutritious item. For example, in one environment nutritious items in summer are those with a $-1$ in the 0th bit and in winter nutritious items are those with a $1$ in the 1st bit. Anything that is not nutritious is poisonous. Thus, for a given season the ANNs only have to learn which of the input bits is important. In our ANN visualizations, the input nodes that correspond to the decision bits are denoted with a `D' inside the input node.

This work introduces diffusion-based neuromodulation and compares it to standard neuromodulation. As described in the section on~\nameref{sec:learningForgeting}, in standard neuromodulation, regular nodes receive modulatory signals via direct connections from modulatory nodes. In diffusion-based neuromodulation, regular nodes receive modulatory signals based on their location in the ANN and a concentration gradient of modulatory chemical. For the implementation of diffusion-based neuromodulation in this paper, the concentration gradient is produced by two point sources located at the far left and right of the ANN (\ref{S1_Fig}). 

The left and right modulatory point sources are tied to summer and winter feedback respectively. They are high (1) if the previously eaten food item was nutritious, and low (-1) if it was poisonous. The modulatory signals of the left and right point sources remain at 0 in winter and summer respectively (i.e. when not in the season they are informative about), and are 0 if the previously presented food item was not eaten. To save computation, we do not simulate the temporal dynamics of diffusion, but rather assume the diffusion chemicals have already reached a steady state. As soon as the activation of the point source is non-zero the simulated chemical instantaneously fills the space within a radius of 1.5 from the center of the point source, and modulates all of the nodes within that space. The simulated chemical released by a point source does not extend beyond 1.5 units of distance from the point source to model the fact that neurotransmitters in the brain can not diffuse forever, but rather are localized due to various factors such as obstructions in the extracellular space (ECS)~\cite{nicholson1998extracellular}. Lastly, the modulatory signal decreases with distance from a point source. The full implementation details of the ANNs, standard neuromodulation, and diffusion-based neuromodulation can be found in \nameref{sec:Methods}.  

\subsubsection* {Evolutionary Algorithm}

This paper has 4 treatments. Two treatments are from Ellefsen et al.~\cite{ellefsen2015neural} and are individuals with standard (i.e. non-diffusing) neuromodulation evolved to maximize performance alone (PA) and evolved to both maximize performance and minimize a connection cost (PCC) (i.e. the CCT)~\cite{clune2013originModularity}. The other two treatments are the same except their learning rule is diffusion-based neuromodulation. These diffusion treatments are performance alone with diffusion (PA\_D) and performance with a connection cost and diffusion (PCC\_D). All individuals are evolved with the probabilistic, multi-objective evolutionary algorithm PNSGA~\cite{clune2013originModularity}. 50 independent runs for each treatment were performed to gather a large sample size for analysis. A detailed description of the parameters for the EA can be found in \nameref{sec:Methods}. 

To prevent evolution from hard coding the seasonal associations into individuals, an individual's fitness is averaged over 4 lifetimes, each with a different environment. The 4 environments are randomized after every generation, which randomizes the seasonal associations and the food ordering.

\section*{Results}
\label{sec:results}

\subsection*{Performance}

\begin{figure}[h!]
\includegraphics[width=1\linewidth]{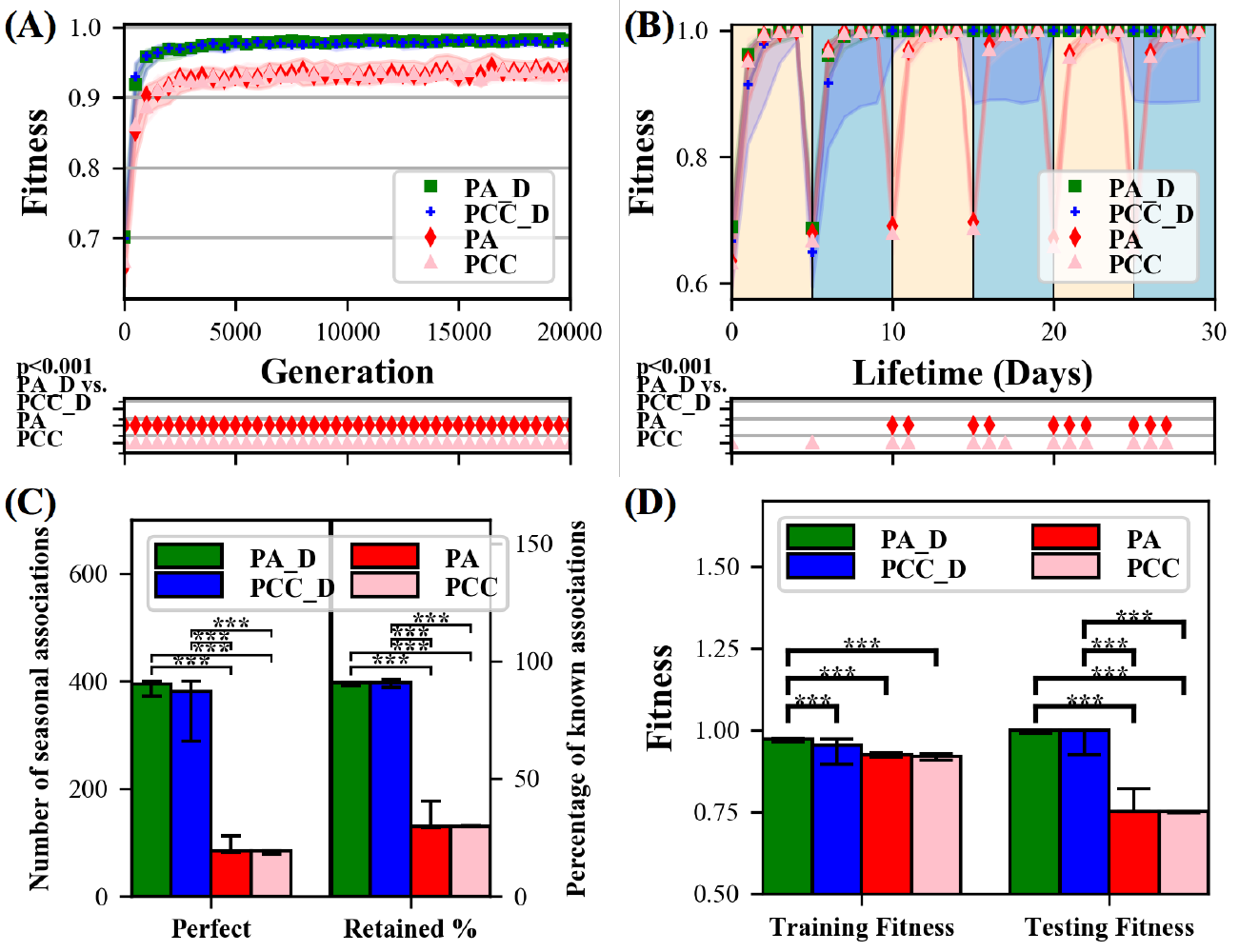} 
 \caption{ \textbf{Diffusion treatments outperform non-diffusion treatments across various metrics.}  \textbf{(A)} Across all generations diffusion treatments ($PA\_D$ $\&$ $PCC\_D$) achieve significantly higher $(p<0.001)$ fitness than non-diffusion ($PA$ $\&$ $PCC$) treatments. \textbf{(B)} Diffusion treatments maintain consistent fitness over their lifetime after the first two seasons, indicating they remember how to solve a task even after they have not performed that task for an entire season. Non-diffusion treatments do not. \textbf{(C)} Diffusion treatments have significantly higher $(p<0.001)$ Retained Percentages and Perfect (i.e. know both summer and winter) seasonal associations than non-diffusion. \textbf{(D)} Diffusion treatments posses significantly higher $(p<0.001)$ testing fitness than non-diffusion treatments. Throughout paper, all statistics are done with the Mann-Whitney U test. Markers below line plots indicate a significant difference ($p<0.001$) between PA\_D and the other treatments at the corresponding data point. For all bar plots, except when stated, a significance bar labeled with `***' is placed between bars that are significant at the level of $p<0.001$. Lastly, the summary value and confidence intervals for all plots in this paper are the median and 75th and 25th percentiles respectively. }
\label{fig:performanceMeasures}
\end{figure}

For all generations, diffusion treatments significantly outperform non-diffusion treatments on the foraging task (Fig~\ref{fig:performanceMeasures}, A). To understand why we performed a post-evolution analysis on the highest fitness individual from the last generation of each evolutionary run. In this analysis, each individual is re-evaluated in 80 new foraging task environments. For each environment, individuals are evaluated first with their initial, evolved weights and learning on (\emph{training phase}), and then again with their learned weights and learning off (\emph{testing phase}). In the training phase, the following metrics (discussed below) are calculated: fitness over lifetime, seasonal associations, training fitness, and weight changes. In the testing phase, the following metrics (discussed below) are calculated: testing fitness and functional modules.

Within a lifetime, diffusion treatments exhibit constant fitness after the first two seasons while the fitness of non-diffusion treatments drops sharply after every season transition (Fig~\ref{fig:performanceMeasures}, B), clearly demonstrating that diffusion treatments have solved catastrophic forgetting on this problem and non-diffusion treatments have not. To complement lifetime fitness, at the end of each season during the training phase individuals are re-evaluated, with learning turned off, to determine what seasonal associations the individual knows. In this re-evaluation, an individual is considered to have \emph{Known} a season's food association if it eats all the nutritious food items and does not eat any poisonous food items for that season. It possesses a \emph{Perfect} seasonal association if it knows the seasonal association for both summer and winter at the end of each season, which tests if the off-season association is still known after training for that season. Aside from random chance, the best an agent can do is have Perfect seasonal associations in 5 of the 6 seasons, because it is not until the end of the second season that they could have learned both sets of season associations (at the end of the first season they have not yet experienced the other season). Summed over 80 environments, the maximum score on the Perfect metric is thus $80\times5=400$. 

Diffusion treatments possess a near-maximum median of 395 ($PA\_D$) and 381 ($PCC\_D$) Perfect associations, respectively (Fig~\ref{fig:performanceMeasures}, C). Both non-diffusion treatments possess a median of 84 Perfect seasonal associations (Fig~\ref{fig:performanceMeasures}, C). These results are further evidence that diffusion treatments, but not non-diffusion treatments, are reliably eliminating catastrophic forgetting. In fact, the only reason the non-diffusing treatments have any Perfect associations is because, due to chance, in 14 of the 80 post-evolution environments the seasonal associations for summer and winter were exactly the same, meaning that learning one seasonal association means both are known. In those instances, it is possible to know both seasonal associations at the end of all 6 seasons without solving catastrophic forgetting, which explains the 84 Perfect seasonal associations for non-diffusion treatments $(14 \times 6=84)$. 

\begin{figure}[!h]
	\centering
         \includegraphics[width=\textwidth]{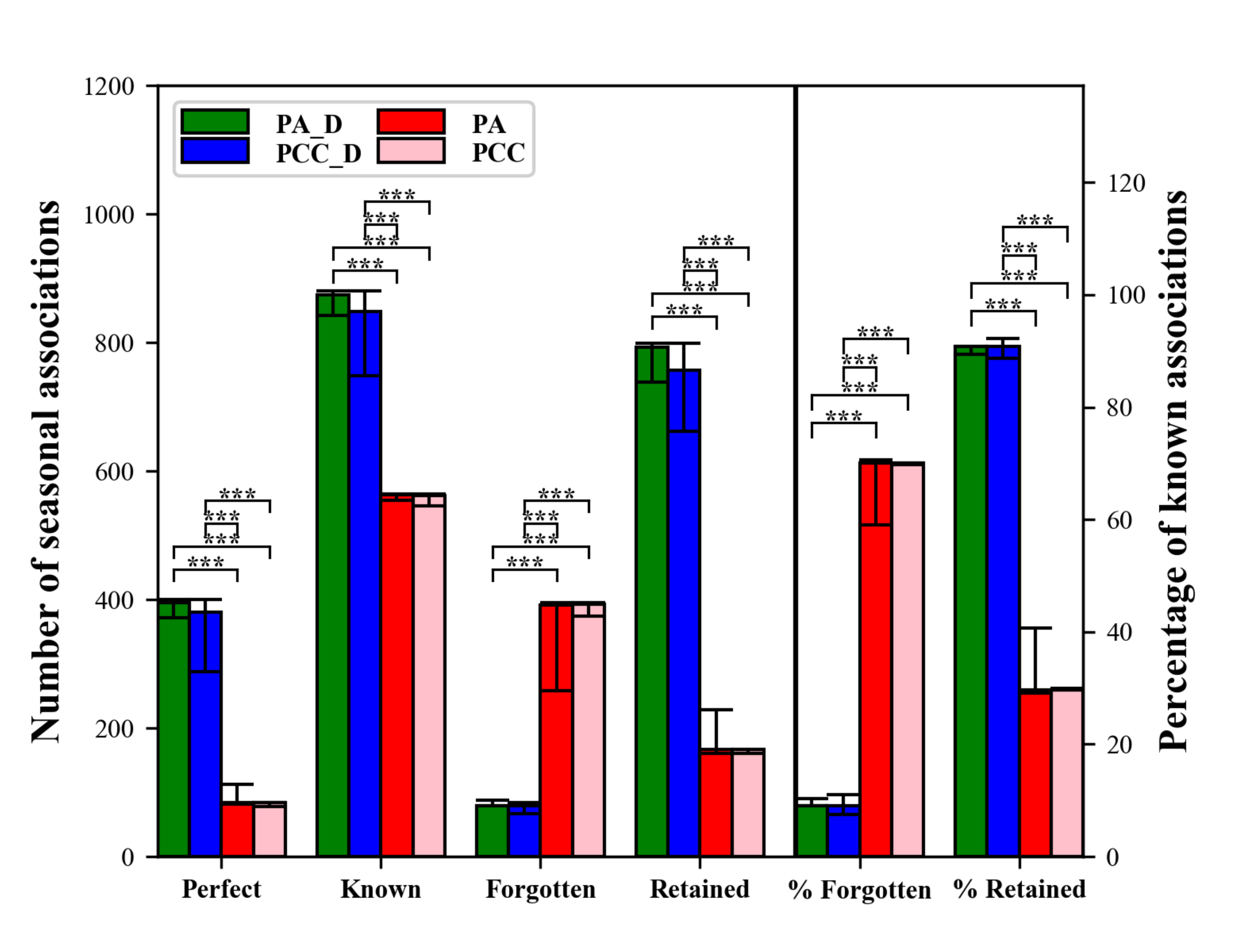}
        \caption{        
        {\bf Plot of all seasonal associations.}  See main text for description and interpretation. For further details on seasonal associations see Ellefsen et al.~\cite{ellefsen2015neural}.
        }
\label{S2_Fig}
\end{figure}

We also calculate how many seasonal associations are \emph{Retained} or \emph{Forgotten} from the prior season. At the end of each season, the maximum number of Known seasonal associations is 2 (i.e. summer and winter), which means that the maximum number of seasonal associations that could have been Retained or Forgotten from the prior season is also 2. The number of Retained seasonal associations is divided by the number of Known seasonal associations to calculate the \emph{Retained Percent} of seasonal associations. See Ellefsen et al.~\cite{ellefsen2015neural} for a more detailed description of seasonal associations and \ref{S2_Fig} for a plot of all seasonal associations. Diffusion treatments have a median Retained Percent of 91.8\% while non-diffusion treatments have a median Retained Percent of 29.6\% (Fig~\ref{fig:performanceMeasures}, C). 

Retained Percent provides an intuitive sense of how many seasonal associations are remembered, but the metric can be misleading since it depends on how many seasonal associations are Known. An individual can achieve a high Retained Percent by not having many Known seasonal associations in the first place. To compliment Retained Percent we calculate the fitness of individuals during the testing phase. If an individual learned and stored information during the training phase then it can do even better during the testing phase because it does not have to make the mistakes inherent in learning. Individuals that have solved catastrophic will have perfect testing fitness. On the other hand, individuals that simply relearn each season will perform poorly during the testing phase because they cannot relearn, and will thus perform well for the last season experienced before the testing phase, not both. Because the base fitness is 0.5 (Eq~\ref{eq:fitnessFunction}), knowing only one of the two seasons results in a testing fitness of 0.75.

Diffusion treatments have a median testing fitness of 1 (Fig~\ref{fig:performanceMeasures}, D), which is an increase from the training fitness, and the max value, revealing that a majority of individuals have learned to solve the tasks perfectly. Non-diffusion treatments exhibit a large decrease from training fitness and end up with a median testing fitness of 0.75 (Fig~\ref{fig:performanceMeasures}, D). This evidence, along with the performance drops after season transitions (Fig~\ref{fig:performanceMeasures}, B) and the low number of Perfect seasonal associations (Fig~\ref{fig:performanceMeasures}, C), confirms that the non-diffusion treatments are continuously forgetting and relearning each season. The original fitness broken down by season, provided for comparison in a knockout analysis for the functional modules discussed in the next section (\ref{S3_Fig}), confirms that only the last season seen in the training phase (the winter season) is known after the training phase in non-diffusion treatments.

In conclusion, fitness over lifetime, the number of Perfect seasonal associations, Retained Percent, and testing fitness all indicate that a majority of the individuals in the diffusion treatments are solving catastrophic while individuals in the non-diffusion treatments are not. 

\subsection*{Functional Modules}

The main idea of this work and Ellefsen et al.~\cite{ellefsen2015neural} is that the isolation of information in functional modules could help reduce interference and mitigate catastrophic forgetting. To identify functional modules within ANNs we introduce the activation record knockout (ARK) algorithm (See~\nameref{sec:Methods}). ARK is based on the subsets regression on network connectivity (SRC) algorithm~\cite{velez2016identifying}. Like SRC, ARK can identify the nodes and connections responsible for specific subproblems in order to identify functional modules within an ANN. The end result is a core functional network (CFN), which is a subnetwork of the ANN that possesses at least the same fitness as the original network. ARK is applied to the final learned networks at the end of each training phase, and is based on the node activations gathered (i.e. activation record) during the testing phase. Because each individual is evaluated against 80 different environments, each with their own training and testing phase, there are 80 different CFNs for each individual.

\begin{figure}[h!]
\includegraphics[width=\textwidth]{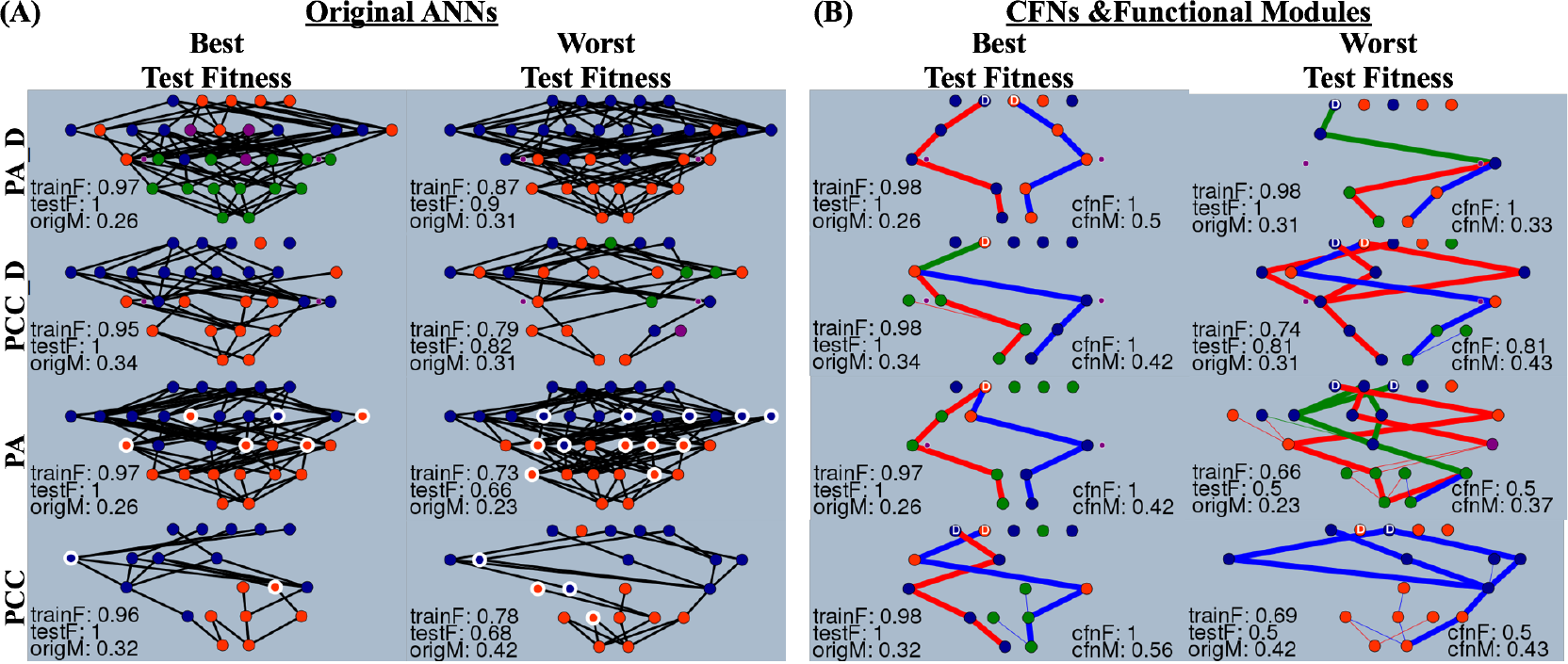}
\caption{\textbf{High-performing networks are differentiated from low-performing networks through the presence of distinct functional modules in Core Functional Networks (CFNs), network features not seen when just examining the original, non-simplified ANNs. } \textbf{(A)} Original ANNs for the networks with the best and worst test fitness. Inset text is the training fitness (trainF), testing fitness (testF), and structural modularity of the original ANN (origM) averaged over all 80 environments in the post-evolution analysis. Superficially there is nothing that distinguishes networks that have the best testing fitness. \textbf{(B)} One example CFN for each of the corresponding ANN from A. Inset text is the structural modularity of the original ANN (origM), training fitness (trainF), testing fitness (testF), CFN fitness (cfnF), and CFN modularity (cfnM) for the environment that produced the CFN. High-performing networks possess sparse CFNs with either two distinct functional modules (red and blue) that form separate paths, or a common functional module (green) that branches off into two distinct functional modules. Low-performing networks possess CFNs that are much more entangled, or do not connect to the decision input bits (input nodes marked with `D') or season outputs. Structural modularity is quantified with the Q-Score metric~\cite{Leicht2008}. 20 additional CFNs for the best and worst individuals are provided in \ref{S4_Fig}, \ref{S5_Fig}, \ref{S6_Fig}, and \ref{S7_Fig}. For diffusion ANNs the locations of the point sources are indicated by small, purple, filled circles (\ref{S1_Fig}) and the modulatory nodes for non-diffusion ANNs are indicated by circles with thick white borders. Nodes whose activation variance is below $1.0\times10^{-9}$ are deemed to be bias nodes and are visualized with thin, outgoing connections. }
\label{fig:bestWorstAdultFitness}
\end{figure}

For the ANNs with the highest and lowest testing fitness for each treatment, Fig~\ref{fig:bestWorstAdultFitness} shows the original (non-simplified) networks, an example CFN, and its functional modules. For the foraging task, we define three types of functional modules: connections that encode for the summer task, connections that encode for the winter task, and connections that are in common and encode for both season tasks. These connections are colored red, blue, and green respectively in the CFN visualizations. See \nameref{sec:Methods} for details on how ARK identifies functional modules. 

\begin{figure}[h!]
	\centering
         \includegraphics[width=\textwidth]{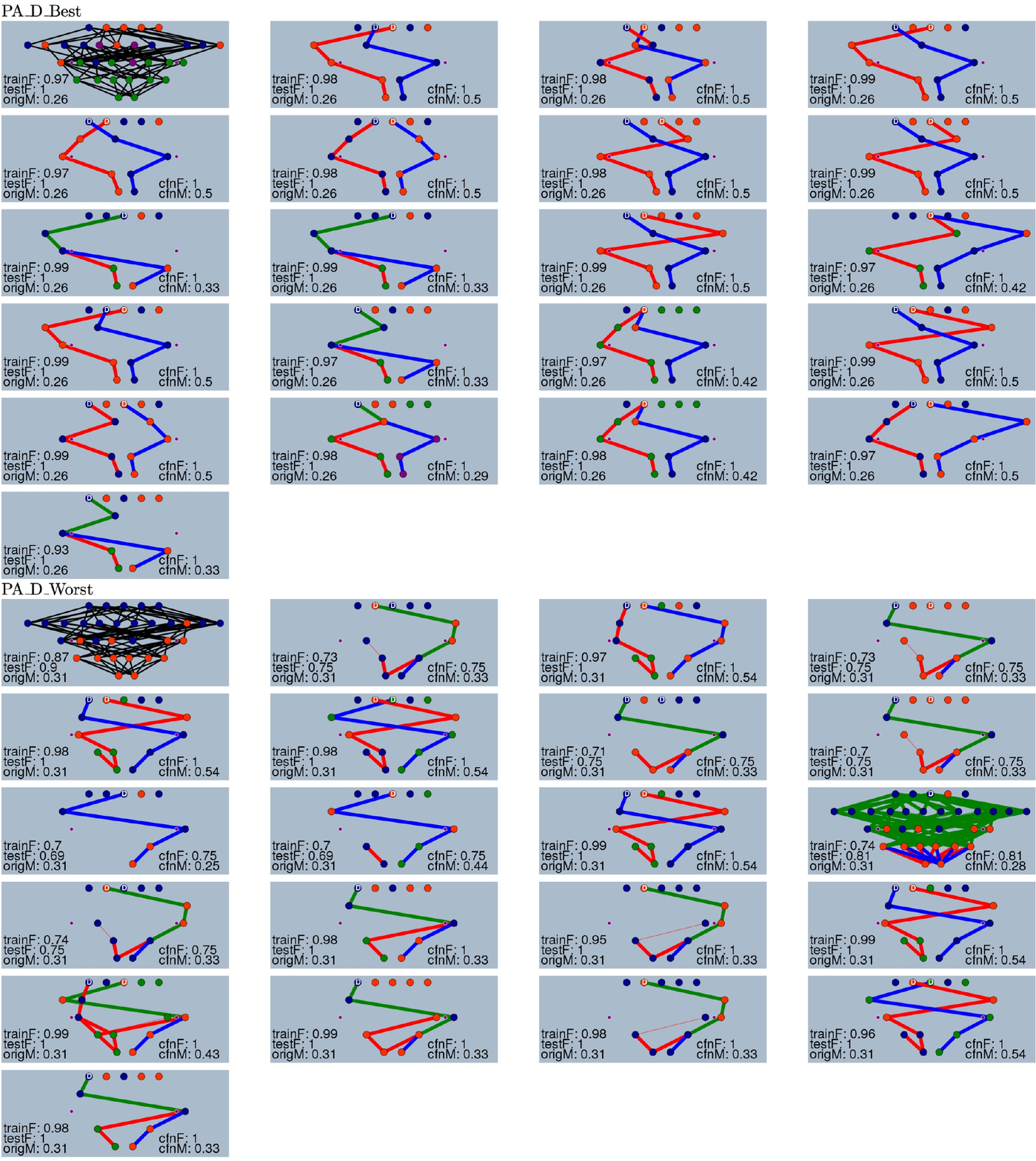}
         \caption{        
{\bf 20 random CFNs for the best and worst PA\_D individuals.} Each block contains the unsimplified version of the individual followed by 20 of its CFNs. }
\label{S4_Fig}
\end{figure}

\begin{figure}[h!]
	\centering
         \includegraphics[width=\textwidth]{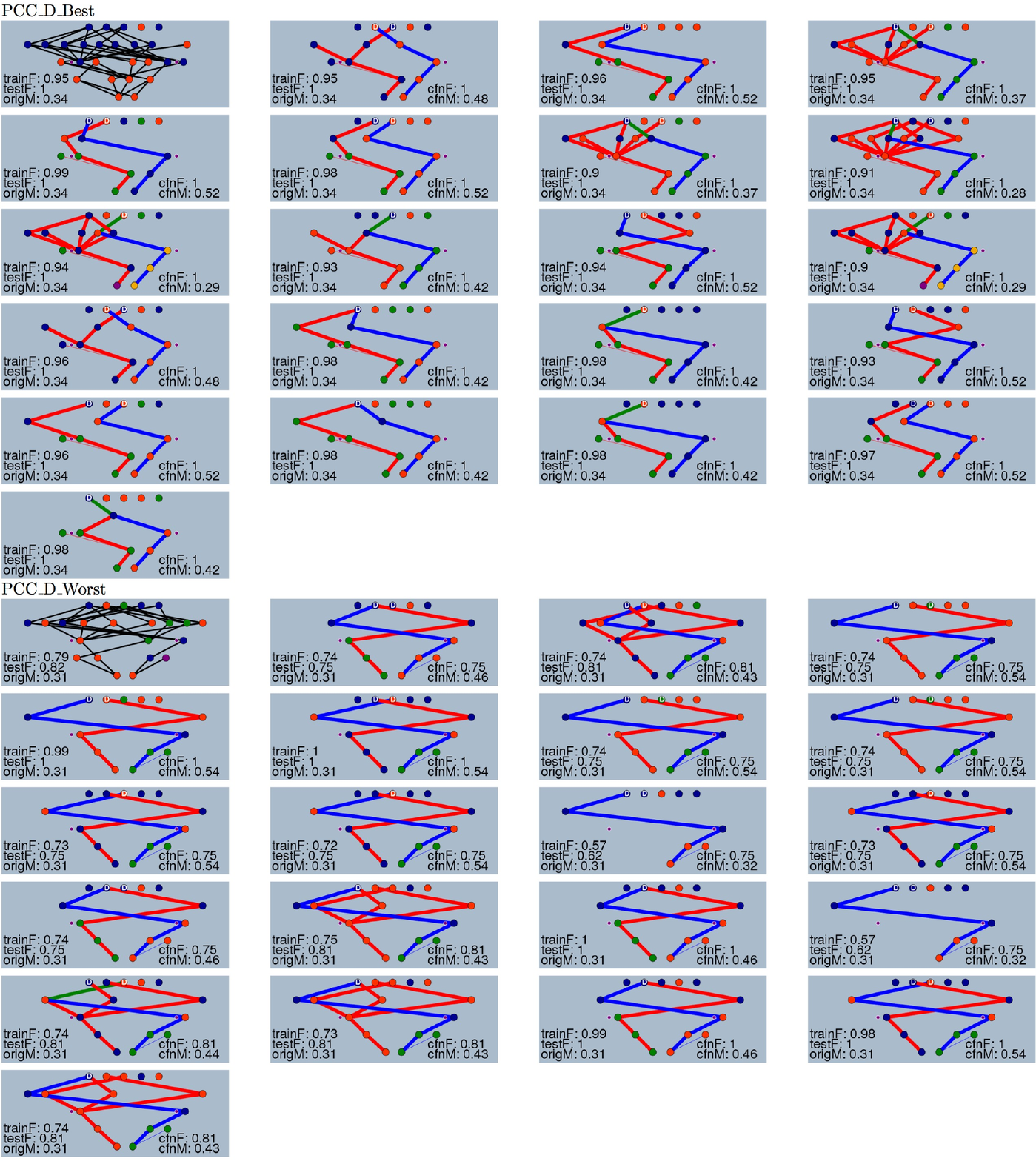}
         \caption{        
{\bf 20 random CFNs for the best and worst PCC\_D individuals.} Each block contains the unsimplified version of the individual followed by 20 of its CFNs. }
\label{S5_Fig}
\end{figure}

\begin{figure}[h!]
	\centering
	 \includegraphics[width=\textwidth]{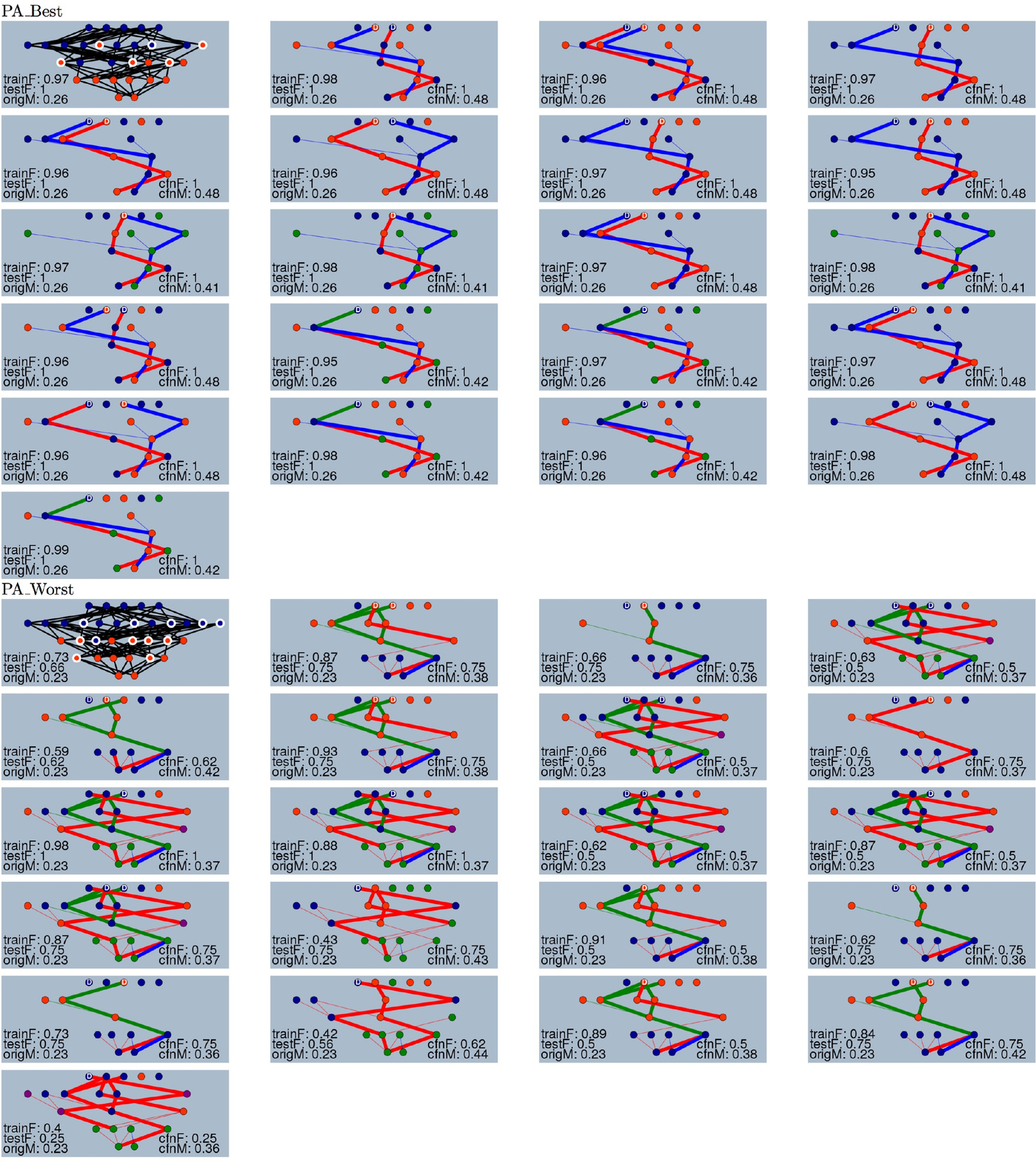}
           \caption{        
{\bf 20 random CFNs for the best and worst PA individuals.} Each block contains the unsimplified version of the individual followed by 20 of its CFNs. }
\label{S6_Fig}
\end{figure}

\begin{figure}[h!]
	\centering
	 \includegraphics[width=\textwidth]{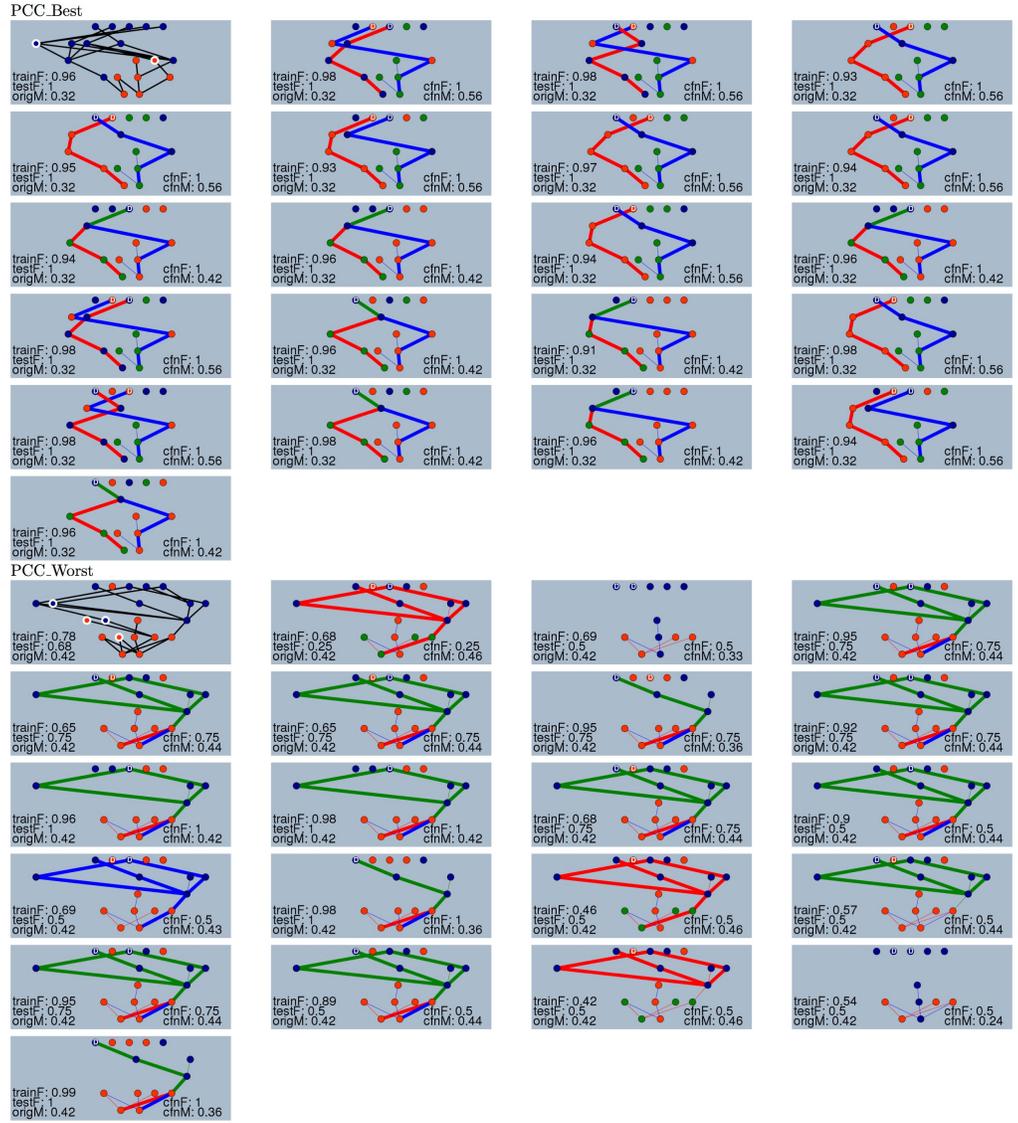}
        \caption{        
{\bf 20 random CFNs for the best and worst PCC individuals.} Each block contains the unsimplified version of the individual followed by 20 of its CFNs. }
\label{S7_Fig}
\end{figure}

The following descriptions of the CFNs in this work are qualitative, but provide a sense of how information is encoded and processed in the networks. The majority of CFNs for the highest performing diffusion and non-diffusion individuals come in two variants (Fig~\ref{fig:bestWorstAdultFitness}, B). The first, and most predominant, possesses two separate functional modules, one for summer and one for winter, that connect the decision bit inputs for summer and winter to the summer and winter output. The second, which can occur when the decision bit input is the same for both seasons, possesses a single common connection from the decision bit input that then branches into two separate functional modules. In both of these instances, there is no interference with the seasonal information as it progresses through the network. The CFNs for the lowest performing, non-diffusion individuals exhibit many patterns, but in general possess two common themes (Fig~\ref{fig:bestWorstAdultFitness}, B). The first is that a decision bit input or season output is not part of any functional module, and is disconnected from the CFN. The second is that there are connections from unimportant inputs or laterally between functional modules. The first pattern prevents the CFN from receiving or transmitting season specific information and the second pattern produces interference as seasonal information progresses through the network. The CFNs for the lowest performing diffusion individuals, whose testing fitness is mid-range between the best and worst testing fitness values across all treatments, possess a mix of the low and high-performing CFN patterns. Thus, it is visually apparent that the low-performing CFNs are slightly less modular and sparse than high-performing CFNs (Fig~\ref{fig:bestWorstAdultFitness}, B). A larger sample of CFNs for the best and worst individuals is provided in \ref{S4_Fig}, \ref{S5_Fig}, \ref{S6_Fig}, and \ref{S7_Fig}.

\begin{figure}[h]
\includegraphics[width=1\linewidth]{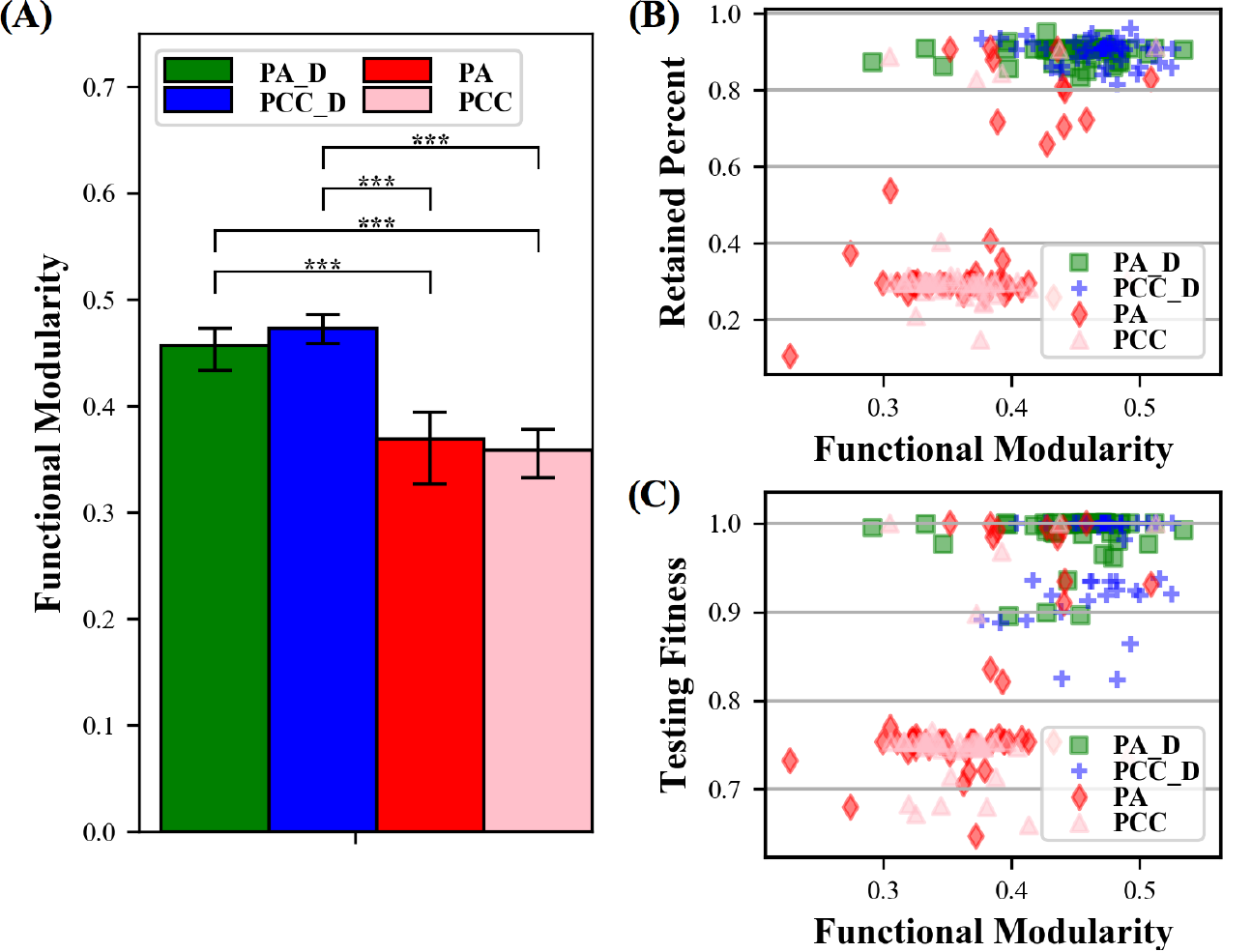} 
\caption{\textbf{Structural modularity scores for Core Functional Networks (CFNs) (i.e. functional modularity) sets diffusion networks apart from non-diffusion treatments.} \textbf{(A)} Structural modularity Q-Scores for Core Functional Networks (CFNs) (i.e. functional modularity). \textbf{(B,C)} Scatter plots of functional modularity versus two quantifiable measures of high performance: Retained Percent and testing fitness.}
\label{fig:cfnModularity}
\end{figure}

The structural modularity of the CFNs (i.e. functional modularity) reflects the patterns described above and reveals a clear difference between the diffusion treatments, which are predominantly high-performing, and the non-diffusion treatments, which are predominantly low-performing (Fig~\ref{fig:cfnModularity}, A). The identification of the CFNs and functional modules reveal two insights. The first is that diffusion-based neuromodulation can be a strong inducer of functional modularity (Fig~\ref{fig:cfnModularity}, A). Second, if the CFN of an ANN is highly modular, regardless of diffusion, it will exhibit less, or no, catastrophic forgetting (Fig~\ref{fig:cfnModularity}, B and C).  In the upper right quadrant of the scatter plots in Fig~\ref{fig:cfnModularity} (B and C), where high-performing, high functional modularity ANNs are placed, there are mostly diffusion networks, but there are also a few non-diffusion networks. Thus, diffusion-based neuromodulation is a strong inducer of functional modules, but it is the functional modules that are allowing catastrophic forgetting to be mitigated.

\section*{Task-Specific Localized Learning}
\begin{figure}[h!]
	\centering
         \includegraphics[width=\textwidth]{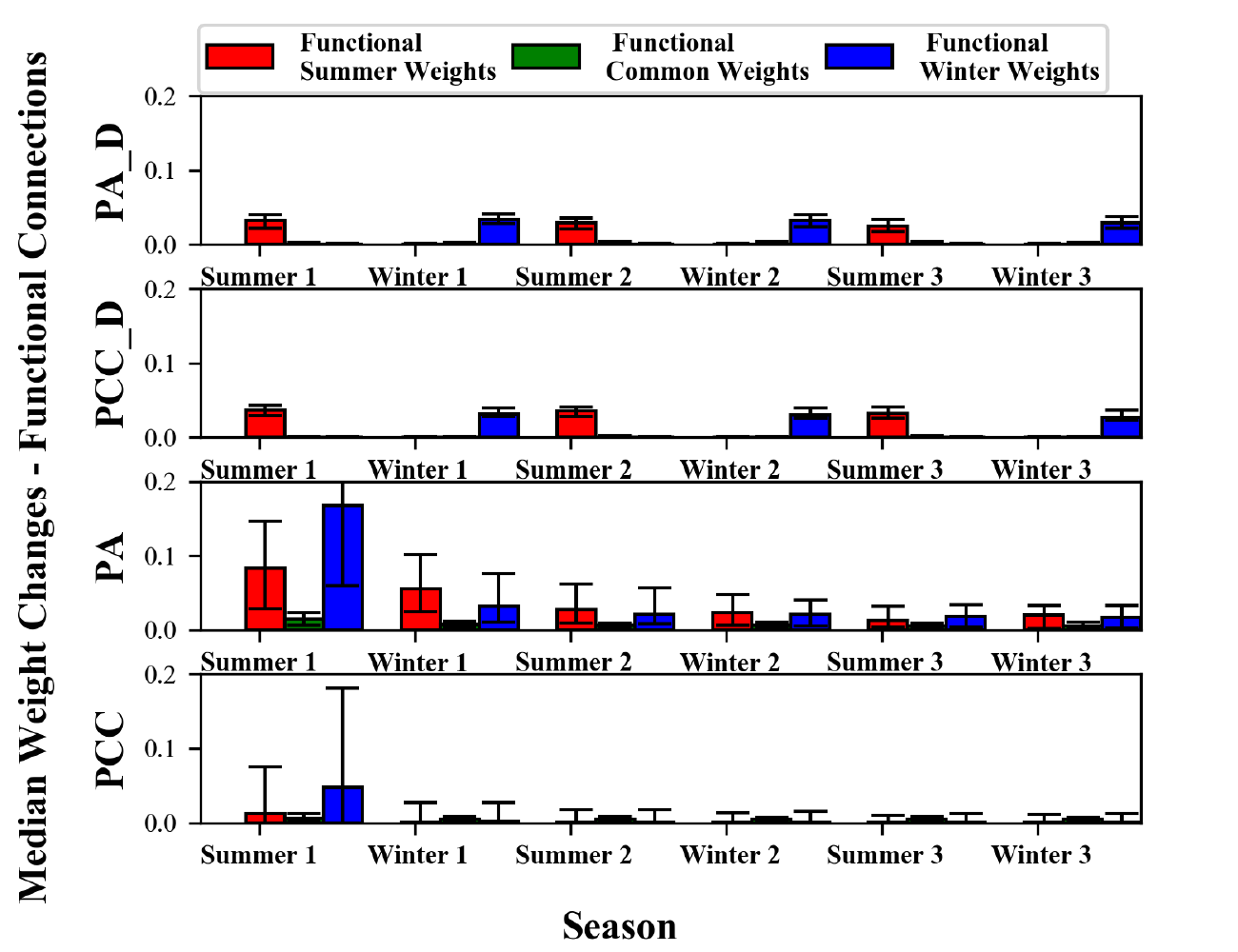}
	\hspace{-10pt}
	\vspace{0pt}
        \caption{\textbf{Diffusion treatments change only connections that will become the summer and winter functional modules in those respective seasons, while non-diffusion treatments change either all connections $(PA)$ or only common connections every season $(PCC)$.} Note, weight change is also occurring in the other connections within range of the points sources, but we plot only connections that eventually become functional and encode task information. Bars are not statistically compared to one another.
        }
\label{fig:weightChange}
\end{figure}

Our hypothesis is that diffusion-based neuromodulation produces the functional modules shown in the prior section by inducing task-specific learning in a specific group of nodes and connections. While such coordination is theoretically possible in non-diffusion networks, we hypothesized that it would be less likely because it requires many separate mutations to create individual connections that produce a coordinated effect. To investigate whether such task-specific or coordinated learning occurs in diffusion or non-diffusion treatments, we record and plot median weight changes for connections that will become the functional modules (Fig~\ref{fig:weightChange}). 

For the diffusion treatments, learning in a given season is isolated to specific groups of nodes and connections (Fig~\ref{fig:weightChange}). During summer, the connections that undergo weight changes are those that will form the summer functional module. During winter, a \emph{different} group of connections, those that will form the winter functional module, undergo weight changes. This task-specific localized learning eliminates catastrophic forgetting. Within each season, learning is turned on and off in a specific group of connections, leaving nodes and connections in the rest of the ANN, and whatever seasonal information they encode, alone and intact. In contrast, in non-diffusion ANNs learning is not task-specific (Fig~\ref{fig:weightChange}), and weight changes occur in connections that will correspond to both seasons. For instance, in winter, non-diffusion treatments experience weight changes in connections that will become responsible for winter, but also in connections that will become responsible for summer, or both. Such interference explains why these treatments exhibit catastrophic forgetting. 

\section*{Discussion}

The functional modules in this paper were produced by task-specific localized learning, but in Ellefsen et al.~\cite{ellefsen2015neural} it was hypothesized that the modularity of an ANN should produce functional modules by facilitating \emph{modular learning}. The concepts of task-specific localized learning and modular learning are similar in that they induce task-specific learning in specific groups of nodes and connections. The difference is that in modular learning the weight change is induced in a module, while in task-specific localized learning the weight change is induced within a spatial region of the ANN that may or may not be modular. While task-specific localized learning initiates the process in the experiments in this paper, an argument could be made that modular learning is also occurring. At some point during task-specific localized learning, a functional module forms and subsequent learning occurs within it. It is also possible that evolution sets the stage for the functional modules that emerge during the localized learning. Investigating the extent to which either mechanism occurs is beyond the scope of this paper, but is an interesting opportunity for future research.

Previous research has shown that a connection cost can improve performance and evolvability~\cite{clune2013originModularity,mengistu2016evolutionary}, and Ellefsen et al.~\cite{ellefsen2015neural} specifically showed that on a foraging task similar to the one in this paper. In this work, we do not see a performance difference between a connection cost (PCC) and not having one (PA). The likely reason is because the problem in this paper is easier than that in Ellefsen et al.~\cite{ellefsen2015neural} such that PA performs well enough without the extra performance boost typically provided by a connection cost. We made the problem simpler and more modularly decomposable in order to better encourage the discovery of functional modularity and investigate whether it can aid with catastrophic forgetting.

The task-specific localized learning in this paper was produced by a new learning algorithm for ANNs called diffusion-based neuromodulation. Diffusion-based neuromodulation is a form of volume transmission because nodes receive information not from direct connections from other nodes, but based on their location within an ANN and a concentration gradient of signaling chemical. Volume transmission can induce elements of modularity as shown by the functional modules in this work, and the structural modularity of GasNets~\cite{fine2006spatially}. Volume transmission, either via a learning signal or an activation signal, could also influence other structural qualities such as regularity~\cite{lipson2007principles,huizinga2014evolving} and hierarchy~\cite{mengistu2016evolutionary}. Regularity means the same connectivity patterns are reused in an ANN. The effect of those repeated connectivity patterns is that large groups of nodes receive the same signal. Volume transmission could produce a similar effect by releasing a chemical signal that can diffuse and excite or inhibit a large group of nodes simultaneously. Lastly, volume transmission could also induce elements of hierarchy as shown by the work on diffusion-limited aggregation and its ability to produce fractal-like patterns known as Brownian trees~\cite{witten1981diffusion}.

The goal of this work is to investigate whether the addition of diffusion to neuromodulation produces functional modules, and if these functional modules would aid in the mitigation of catastrophic forgetting. To accomplish this goal we designed a simple modular forgetting task where we knew the modular decomposition a priori in order to test whether either treatment would discover it. We also designed the implementation of the diffusion-based neuromodulation to best encourage the expected optimal, modular solution to the problem, in order to see whether diffusion helps in the case we most expect it should. That included having the concentration gradient of the modulatory chemical be produced by two points sources tied to the feedback for the tasks in the multitask problem. These point sources are a simple way to specify a concentration gradient, but require the experimenter to specify the number, location, and modulatory signal of these point sources. In future work we will explore how this new approach can scale to much harder, less hand-designed, problems. One such path is to remove the hand-placed point sources and evolve the location and connectivity of modulatory nodes that can release a diffusing modulatory chemical. GasNets evolve the location, connectivity, and diffusion parameters for diffusing nodes~\cite{husbands1998better}. In preliminary experiments for this paper, we found that it was difficult for evolution to specify the location and connectivity of diffusing, modulatory nodes. Evolution would cause many erroneous connections to be fed into the modulatory nodes, which prevented learning from being task-specific, or modulatory nodes were too close to each other, which prevent learning from being localized.  Future work is required to return to the question of whether and how well evolution can place diffusing modulatory nodes, or simplified point sources. Future work could also investigate other methods to produce a modulatory concentration gradient. One option is to specify the concentration gradient with a compositional pattern producing network (CPPN)~\cite{stanley2007compositional}. CPPNs can abstract the concentration gradients of morphogens in order to produce regular patterns of expression. A CPPN could be evolved to produce a concentration gradient of modulatory chemical for every point within an ANN. A prior paper on neuromodulation has already shown that a CPPN can specify the learning rule for connections (i.e. parameters for Hebbian or neuromodulation learning) based on their geometric positions within an ANN~\cite{risi2010indirectly}. 

Our research strategy resembles recent, exciting work on catastrophic forgetting by another research group. They too first hand-coded elements of a DNN's modularity in order to investigate whether modular DNNs are less susceptible to catastrophic forgetting when combined with learning being selectively turned off and on for different tasks. They accomplished the latter by freezing the weights in a module that learned an initial task and allowed learning to occur in a second module that could leverage features from the first module~\cite{rusu2016progressive}. In follow-up work they harnessed these insights to develop a more automated, elegant, less hand-designed method~\cite{Kirkpatrick14032017}, which we also envision is possible with diffusion-based neuromodulation.

\section*{Conclusion}

Catastrophic forgetting is a major challenge that hinders our ability to produce ANNs and general AI that can learn and refine a multitude of different skills and abilities over a lifetime. Ellefsen et al.~\cite{ellefsen2015neural} proposed that the isolation of task-specific information to functional modules should help mitigate catastrophic forgetting. To produce functional modules Ellefsen et al.~\cite{ellefsen2015neural} evolved modular ANNs, via a connection cost, because that would allow for modular learning; where task-specific learning is turned on and off in different modules. While Ellefsen et al.~\cite{ellefsen2015neural} showed that modular ANNs suffered less from catastrophic forgetting they did not see the emergence of different modules for different tasks, or the complete avoidance of catastrophic forgetting. In this paper, we have presented diffusion-based neuromodulation and shown that functional modules for the different tasks appear when task-specific learning occurs in a local group of nodes and connections (i.e. task-specific localized learning). In our experiments, such task-specific localized learning results in the complete avoidance of catastrophic forgetting. This paper thus confirms the central hypothesis of Ellefsen et al.~\cite{ellefsen2015neural}, which is that localized, task-specific learning can form functional modules and solve catastrophic forgetting. Of course, here we have only shown that on a simple problem and simple ANNs. Future work is needed to test the ability of this mechanism to scale to far more challenging problems and larger neural networks. 

\section*{Materials and Methods}
\label{sec:Methods}

The experiment details are adapted from Ellefsen et al.~\cite{ellefsen2015neural}, which is based in the Sferes 2 framework~\cite{mouret2010sferes}. Neuromodulation is modeled off the work of Soltoggio et al.~\cite{soltoggio2008evolutionary}, and was adapted for Sferes 2 by Tonelli and Mouret~\cite{tonelli2011relationships}. Network and EA implementation details follow from prior work with Sferes 2~\cite{clune2013originModularity,mengistu2016evolutionary,huizinga2014evolving}. The software to reproduce these experiments and analyze the data, as well as the key data from our experiments, can be found at https://doi.org/10.15786/M21G6W.

\subsection*{Network Activation}

For all ANNs in this paper, the activation $a_i$ of node $i$ is determined by Eq~\ref{eq:nodeActivation} and Eq~\ref{eq:actFunction}, where $w_{ij}$ is the weight from node $j$ to node $i$, $b_i$ is the internal bias of node $i$, and $C_n$ are all non-modulatory nodes with direct connections to node $i$.

\begin{equation}
\label{eq:nodeActivation}
a_{i}=\phi \left ( \sum_{j\epsilon C_{n}}w_{ij}a_{j}+b_{i} \right )   
\end{equation}

\begin{equation}
\label{eq:actFunction}
\phi(x)=\frac{2}{1+e^{-32x}}-1
\end{equation}

The relatively high number of 32 in Eq~\ref{eq:actFunction} makes the transition in the sigmoid function steep and behave more like a step function. 

\subsection*{Learning Rules}
\label{sec:learningRules}

For both diffusion-based neuromodulation and standard neuromodulation, the change in a connection weight $w_{ij}$ is determined by the activation of the two nodes it connects, $a_j$ and $a_i$, a learning rate $\eta$ (0.002 for all experiments), and an external modulatory signal $m_i$ (Eq~\ref{eq:modWeightChange}). The modulatory signal $m_i$ affects all connections feeding into node $i$, and can accelerate, dampen, or invert learning in those connections. 

\begin{equation}
\label{eq:modWeightChange}
\Delta w_{ij}= \eta \cdot m_i \cdot a_i \cdot a_j
\end{equation}

\begin{figure}[h!]
	\centering
	\includegraphics[width=\textwidth]{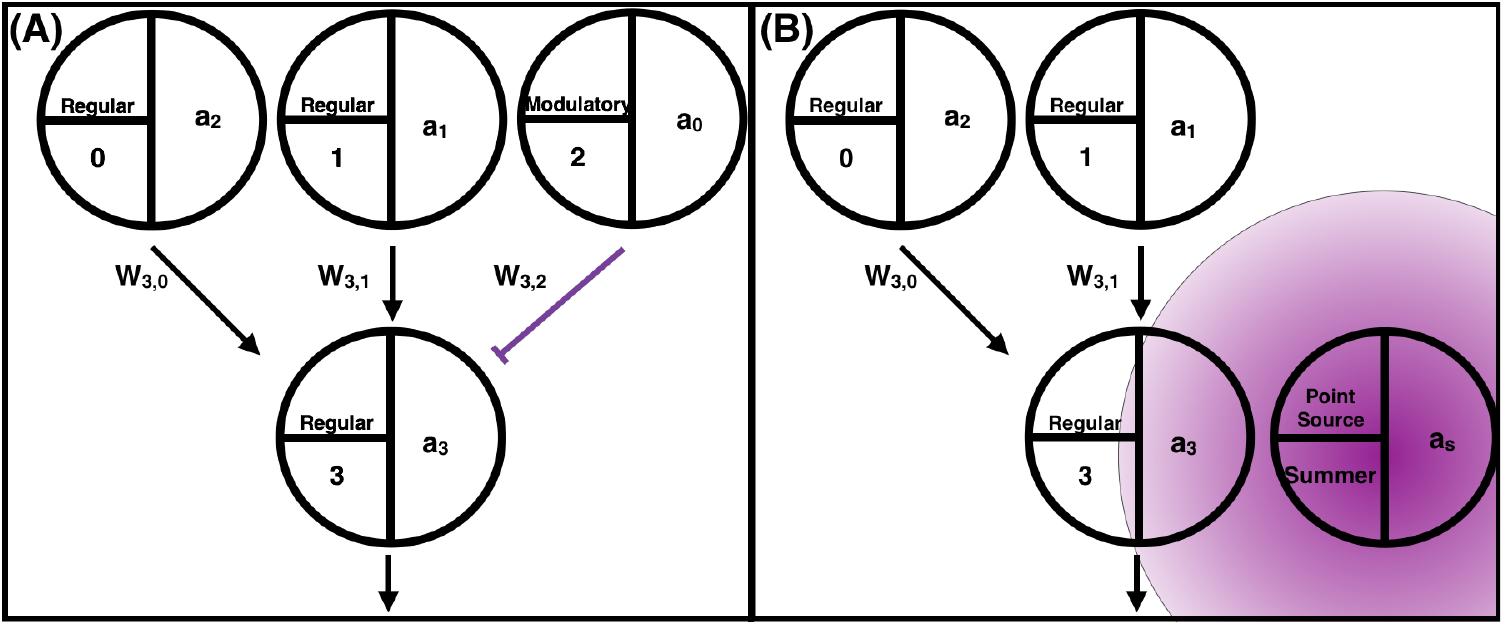}
        \caption{\textbf{An illustration of (A) standard and (B) diffusion-based neuromodulation.} For both, the activation of node $3$ depends on the activations of nodes $0$ and $1$ and the connecting weights $w_{3,0}$ and $w_{3,1}$ (Eq~\ref{eq:actFunction}). The changes in weights $w_{3,0}$ and $w_{3,1}$ rely on the sum of modulatory signals received by node $3$ (Eq~\ref{eq:modWeightChange}). \textbf{(A)} In standard neuromodulation, the modulatory signal comes from the direct connection from the modulatory node $2$ (Eq~\ref{eq:standardModSignal}). \textbf{(B)} In the diffusion-based neuromodulation implementation in this paper, the modulatory signal comes from the concentration gradients released by the point sources. In this example, the modulatory signal of node 3 is determined by its distance to the summer point source.}
\label{fig:nmFigure}
\end{figure}

For a node $i$, in standard neuromodulation, the modulatory factor $m_i$ in Eq~\ref{eq:modWeightChange} is obtained by summing up the activations transmitted to node $i$ through connections originating from modulatory nodes ($C_m$) (Eq~\ref{eq:standardModSignal}) (Fig~\ref{fig:nmFigure}). For diffusion-based neuromodulation, there are no modulatory nodes. The modulatory factor $m_i$ of node $i$ depends on the activation, $a_s$ and $a_w$, of the summer and winter point sources (Eq~\ref{eq:diffusionModSignal}), and the node's distance, $d_{is}$ and $d_{iw}$, from the summer and winter point sources. The summer point source is located at (-3,2) (\ref{S1_Fig}), and its activation $a_s$ is the feedback for the summer season. The winter point source is located at (3,2), and its activation $a_w$ is the feedback for the winter season. If a node is within 1.5 units of distance from a point source, the strength of the modulatory signal rises according to a Gaussian function as it gets closer to the source (Eq~\ref{eq:gaussEquation}), where $\sigma$ is 0.5. If the distance of the node is greater than 1.5 units its modulatory signal is 0. 

\begin{equation}
\label{eq:standardModSignal}
m_{i}=\phi \left ( \sum_{j\epsilon C_{m}} w_{ij} a_{j} \right )
\end{equation}

\begin{equation}
\label{eq:diffusionModSignal}
m_i=\phi\left ( a_sg(d_{is})+a_wg(d_{iw}) \right )
\end{equation}

\begin{equation}
\label{eq:gaussEquation}
g(x)=\begin{cases}
\frac{e^{-2}}{\sqrt{2\sigma^2\pi}}e^{\frac{-x^2}{2\sigma^2}} & \text{ if } x<=1.5 \\ 
0 & \text{ otherwise } 
\end{cases}
\end{equation}

\subsection*{Evolutionary Algorithm}
\label{sec:ea}

All ANNs are evolved with the probabilistic, multi-objective evolutionary algorithm PNSGA~\cite{clune2013originModularity}. PNSGA is an extension of the multi-objective algorithm NSGA-II ~\cite{deb2002fast}. In PNSGA each objective is given a probability that determines how frequently that objective factors into the selection. For all treatments, both performance and behavioral diversity~\cite{Mouret2012} (described below) objectives have a probability of 100\%, while the connection cost objective has a probability of 75\%. The lower probability for connection cost follows from Ellefsen et al.~\cite{ellefsen2015neural}, and represents the notion that a connection cost is likely to be weaker than other selection pressures in nature. The population size of the EA is 400 and it runs for 20000 generations. 50 independent runs were done for each treatment.

The behavior of each individual is represented by a vector, and for each food item that is presented to the individual a 1 or 0 is appended to the behavioral vector depending on whether the individual ate or not. At the end of the individual's lifetime, the average Hamming distance between its behavioral vector and the behavioral vector of every other individual in the population is calculated to produce a behavioral diversity score. Individuals whose behavior (i.e. sequence of eat or not eat actions) is more different from the behavior of others in the population get a higher score while individuals whose behavior is similar to others get a lower score. Following Ellefsen et al.~\cite{ellefsen2015neural}, we include behavioral diversity because it helps evolutionary algorithms avoid local optima~\cite{Mouret2012}.

At the start of evolution, all ANNs are fully connected and the initial weights for all connections are drawn uniformly from the range [-1,1]. The initial bias values for nodes are also drawn from the range of [-1,1]. To give evolution control over whether a node is modulatory or not, each node possesses an additional evolved parameter called \emph{modul} that ranges from [0,1]. For non-diffusion treatments, if a node's modul is below 0.4 then it is modulatory. For diffusion treatments, because there are no modulatory nodes, the modul parameter does nothing.

Following Ellefsen et al.~\cite{ellefsen2015neural}, the ANNs undergo mutation only and not crossover. For network connections, the probability to add or remove a connection is $20\%$. Per connection, the probability of reassigning the source or target of a connection from one node to another is $15\%$ and the probability of changing a weight is 2/$n$, where $n$ is the number of connections in the ANN. The probability of changing the bias and modul for each node is $10\%$. Lastly,  changes in connection weights, biases, and moduls all involve \emph{polynomial mutation}~\cite{2001multi}.

\subsection* {Activation Record Knockout (ARK)}
\label{sec:ark}

The Activation Record Knockout (ARK) algorithm is a simplification and analysis tool based on the subsets regression on network connectivity (SRC) algorithm~\cite{velez2016identifying}. It identifies the nodes and connections within an ANN that are responsible for its overall behavior in order to simplify it down to a core functional network (CFN). A core functional network is a subnetwork of the original ANN that possesses at least the same fitness as the original ANN. Aside from identifying the nodes and connections responsible for overall performance (i.e. a CFN), ARK can also identify the functional modules within an ANN by identifying the nodes and connections responsible for particular subproblems. This section focuses on ARK's implementation on the ANNs in this paper. For a broader discussion of how ARK could be implemented, specifically in the identification of functional modules, we refer the reader to the original SRC paper~\cite{velez2016identifying}. 

For this example, we identify the summer functional \emph{subnetwork}, which is combined with the winter functional subnetwork to produce the summer and winter functional modules. Before the main ARK analysis, the activation of all nodes during the testing phase is recorded to create an \emph{activation record}. The ARK analysis consists of three basic steps that repeat: select a current node to analyze, calculate the contribution of all connection combinations feeding into the current node, and then select one of those connection combinations. 

\begin{figure}[h!]
	\centering
         \includegraphics[width=\textwidth]{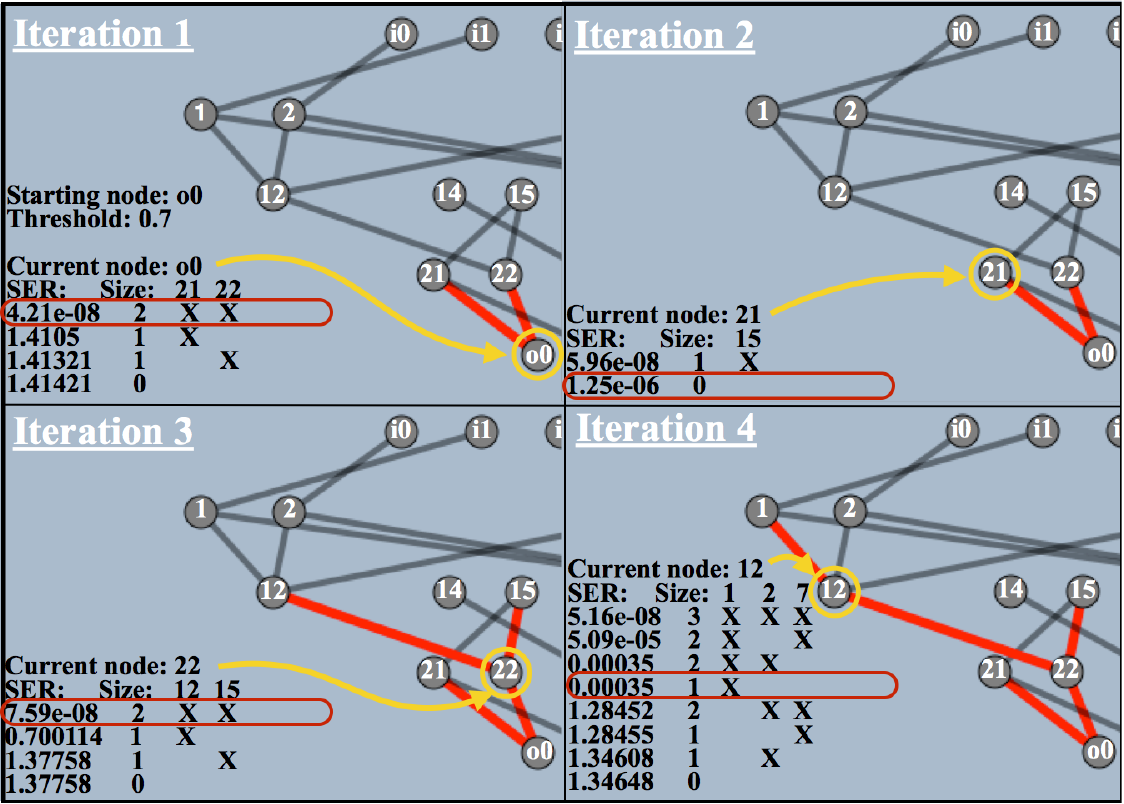}
        \caption{\textbf{The first four steps of the ARK procedure for finding the summer functional network.} }
\label{fig:arkExample}
\end{figure}

To identify the summer functional subnetwork, first, we select the summer output as the current node (Fig~\ref{fig:arkExample}). Second, we perform a $p$-dimensional knockout on all of the connections feeding into the current node, and then compare the resulting knockout activations to the original activation to generate a measure of sensitivity. $p$ is the number of connections feeding into the current node and a $p$-dimensional knockout means we iterate through all (i.e. $2^p$) knockout combinations of incoming connections. We recalculate the activation of the current node given each knockout combination to produce knockout activations. To assess sensitivity, which is the effect on the current node's activation given a combination of its incoming connections, we compute the standard error of regression (SER) between the original activation ($y$) and each knockout activation ($\hat{y}$) (Eq~\ref{eq:ser}). $n$ in Eq~\ref{eq:ser} is the length of the activation record, and the number of different inputs presented to the ANN in a single environment in the testing phase. The ARK table for node o0 (upper left of Fig~\ref{fig:arkExample}, Iteration 1) shows the different knockout combinations for node o0 and their resulting SER values.

\begin{equation}
\label{eq:ser}
SER=\sqrt{\frac{ \sum_{i=0}^{n} (y_i-\hat{y_i})^2} {n}}
\end{equation}

Each current node has its own ARK table that displays the name of the current node and the size and SER for all knockout combinations for that node. For each combination, the connections that are not knocked out are counted towards the size of that combination, and indicated by an `X' in the table. The ARK table for the starting node displays the starting node and error threshold. The error threshold will be discussed shortly, but for all iterations of the ARK procedure in Fig~\ref{fig:arkExample} it is 0.70. Lastly, the combinations are sorted by their SER and in the case of ties reverse sorted by combination size. Note that the no knockout combination (i.e. neither node 21 or 22 is removed) at the top of the ARK table for node o0 (Fig~\ref{fig:arkExample}, Iteration 1) should have a SER of 0, because with no knockout the knockout activation of the current node should be the same as its original activation. The small but non-zero value is due to floating point precision errors and is present in ARK tables for all nodes. It will be discussed in relation to the error threshold.

In the third step of the ARK procedure, we select the smallest combination with a SER less than or equal to the error threshold. The three basic steps of the ARK procedure then repeat with new current nodes selected via breadth first search. Fig~\ref{fig:arkExample} shows 3 more iterations of the ARK algorithm given current nodes 21, 22, and 12. In iteration 2 of Fig~\ref{fig:arkExample}, the empty combination (i.e. no incoming connections) is chosen because it is the smallest combination with a SER below the threshold. The selection of the empty combination suggests that node 21 is acting as a bias node. In iteration 4 of Fig~\ref{fig:arkExample}, the ARK table for node 12 shows that the smallest combination with an SER less than or equal to the error threshold is the one that removes connections from node 2 to node 12 and from node 7 to node 12. ARK stops once there are no more nodes to explore.

\begin{figure}[h!]
	\centering
         \includegraphics[width=\textwidth]{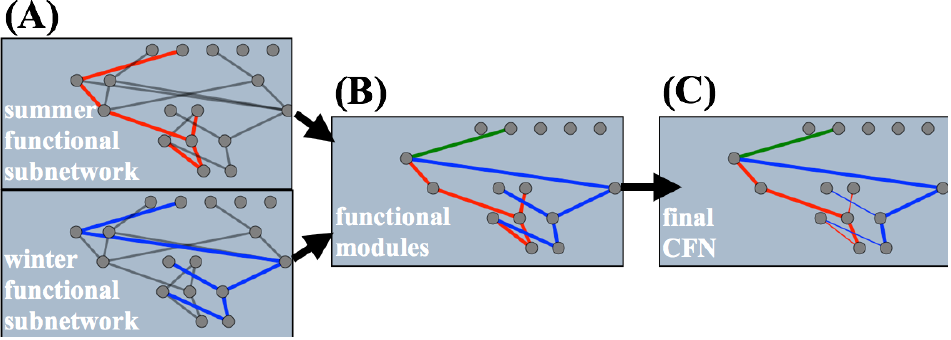}
        \caption{\textbf{Combination of functional subnetworks to produce functional modules.} \textbf{(A)} ARK identifies the functional subnetworks for summer and winter, red and blue connections respectively, in an ANN. \textbf{(B)} Non-functional connections are removed. Functional subnetworks are combined to produce the final functional modules for summer (red connections), winter (blue connections), and common (green connections). \textbf{(C)} As a final visualization technique, connections from bias nodes are made thin.}
\label{fig:combineNets}
\end{figure}

\begin{figure}[h!]
	\centering
         \includegraphics[width=\textwidth]{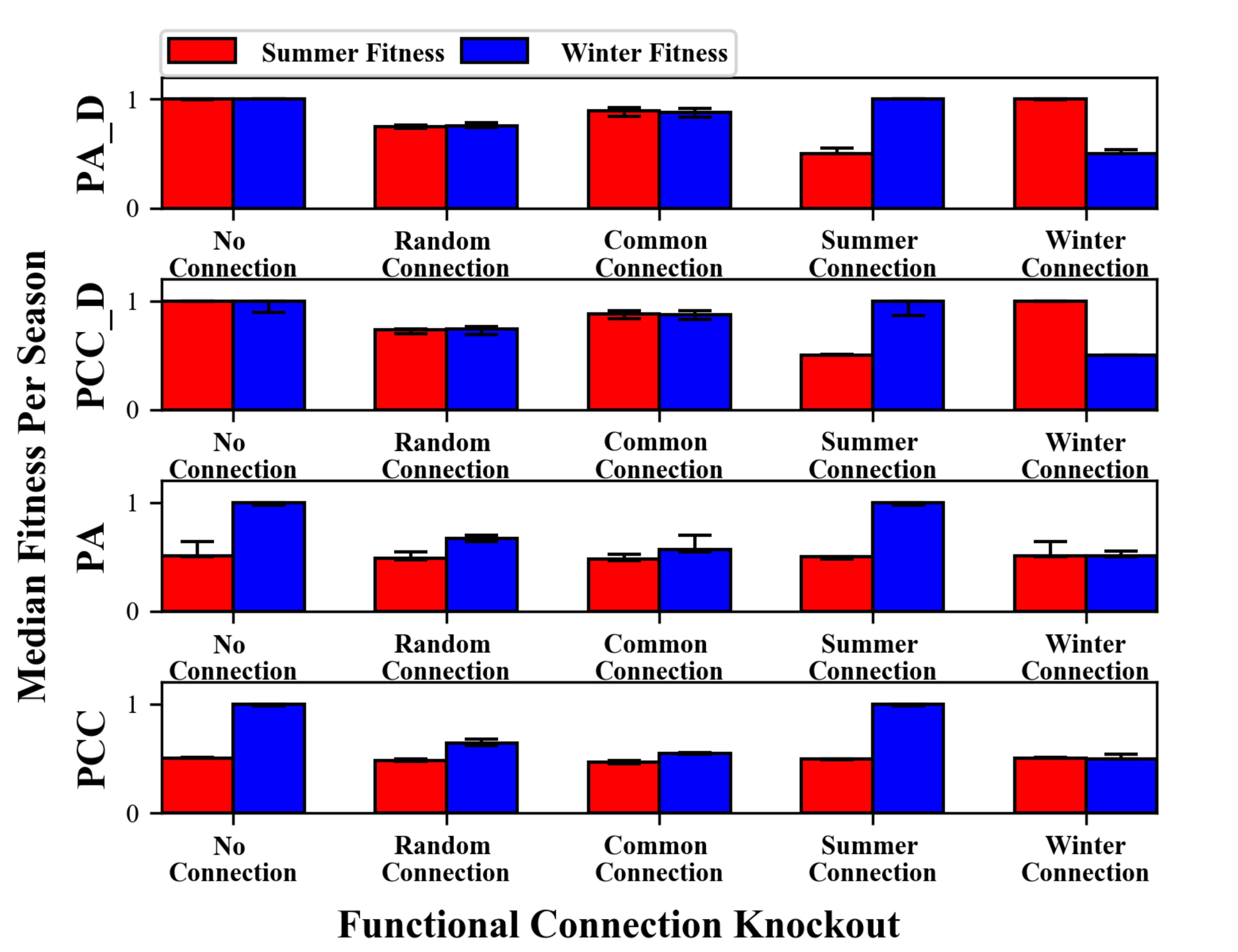}
        \caption{        
{\bf A 1-connection knockout in the Core Functional Networks (CFNs) confirms that ARK properly identifies functional modules.} The original summer and winter fitness for all CFNs is plotted along with the summer and winter fitness after the knockout of a random, common, winter, or summer functional connection. The original (no connection) and random connection fitnesses are provided for comparison. For all treatments, the removal of a summer (or winter) functional connection only causes a drop in summer (or winter) fitness. In contrast, the removal of a common functional connection causes a drop in fitness for both seasons. For non-diffusion treatments, the drop in summer fitness is difficult to see because non-diffusion treatments do not have much competency (i.e. original fitness) on the summer task to begin with. The knockout analysis confirms that the summer and winter functional modules identified by ARK encode for those seasons respectively and that the common functional module identified by ARK encodes for both.}
\label{S3_Fig}
\end{figure}

When ARK stops, the remaining nodes and connections that have not been removed are designated to be the summer functional subnetwork (Fig~\ref{fig:combineNets}, A). To find the winter functional subnetwork the ARK procedure is run again, but this time the start node is o1, which is the output node for the winter season (Fig~\ref{fig:combineNets}, A). Once the winter functional subnetwork is found, it is combined with the summer functional subnetwork to produce a complete picture of the functional modules (Fig~\ref{fig:combineNets}, B). If there are any connections in common between the summer and winter functional subnetworks, then those connections are colored in green and we designate them as a separate \emph{common} functional module that encodes information for both seasons. Following from the prior work on SRC~\cite{velez2016identifying}, a 1-connection knockout is provided in \ref{S3_Fig} that verifies that the functional modules identified by ARK do indeed encode for the summer, winter, or both seasons.

Lastly, separate from the ARK procedure, a variance analysis is done on the activation of all nodes in order to identify possible bias nodes, and gain further understanding of how the ANN works. Any node whose activation has a variance less than $1.0\times10^{-9}$ is deemed to be a bias node and their outgoing connections are made thin in the CFN visualization (Fig~\ref{fig:combineNets}, C). Node 21, discussed previously, is confirmed to be a bias node by the variance analysis.

Each CFN requires an error threshold that determines the aggressiveness of the ARK simplification. Higher thresholds result in the pruning of more connections, but can lead to a loss in fitness. Low thresholds will preserve the fitness of the original ANN, but can result in a lack of simplification and insight. For each CFN, we iteratively increase the error threshold (starting at 0 plus the floating point error) by 0.01, and keep the threshold found right before the CFN starts to lose fitness.

\section*{Acknowledgments}
Jeff Clune and Roby Velez were supported by an NSF CAREER award (CAREER: 1453549). Jeff Clune was also supported by a Wyoming NASA Space Grant Consortium Grant (\#NNX15AI08H) and Roby Velez was supported by a Wyoming Space Consortium Grant (\#NNX10AO95H). All experiments were conducted on the Mount Moran computing cluster at the University of Wyoming Advanced Research Computing Center (ARCC). The authors thank the ARCC staff for their support, and the members of the Evolving AI Lab at the University of Wyoming for valuable feedback on this draft, especially Joost Huizinga, Kai Ellefsen, and Nicholas Cheney. We also thank James Kirkpatrick and one anonymous reviewer for their valuable insights, which improved the manuscript.

\clearpage

\beginsupplement

\section*{Supporting Information}

\vspace{-50pt}

\vspace{-50pt}

\clearpage

\nolinenumbers


\begin{thebibliography}{10}

\bibitem{krizhevsky2012imagenet}
Krizhevsky A, Sutskever I, Hinton GE.
\newblock Imagenet classification with deep convolutional neural networks.
\newblock In: Advances in neural information processing systems; 2012. p.
  1097--1105.

\bibitem{mnih2015human}
Mnih V, Kavukcuoglu K, Silver D, Rusu AA, Veness J, Bellemare MG, et~al.
\newblock Human-level control through deep reinforcement learning.
\newblock Nature. 2015;518(7540):529--533.

\bibitem{levine2016end}
Levine S, Finn C, Darrell T, Abbeel P.
\newblock End-to-end training of deep visuomotor policies.
\newblock Journal of Machine Learning Research. 2016;17(39):1--40.

\bibitem{goodfellow2016deep}
Goodfellow I, Bengio Y, Courville A.
\newblock Deep learning.
\newblock MIT Press; 2016.

\bibitem{french1992semi}
French RM.
\newblock Semi-distributed representations and catastrophic forgetting in
  connectionist networks.
\newblock Connection Science. 1992;4(3-4):365--377.

\bibitem{mermillod2013stability}
Mermillod M, Bugaiska A, Bonin P.
\newblock The stability-plasticity dilemma: investigating the continuum from
  catastrophic forgetting to age-limited learning effects.
\newblock Frontiers in psychology. 2013;4:504.

\bibitem{french1999catastrophic}
French RM.
\newblock Catastrophic forgetting in connectionist networks.
\newblock Trends in cognitive sciences. 1999;3(4):128--135.

\bibitem{Haykin1998}
Haykin S.
\newblock {Neural Networks: A Comprehensive Foundation}.
\newblock 2nd ed. Prentice Hall; 1998.

\bibitem{floreano2008bio}
Floreano D, Mattiussi C.
\newblock Bio-inspired artificial intelligence: theories, methods, and
  technologies.
\newblock The MIT Press; 2008.

\bibitem{haykin2009neural}
Haykin SS.
\newblock Neural networks and learning machines.
\newblock 3rd ed. Prentice Hall; 2009.

\bibitem{widrow199030}
Widrow B, Lehr MA.
\newblock 30 years of adaptive neural networks: perceptron, madaline, and
  backpropagation.
\newblock Proceedings of the IEEE. 1990;78(9):1415--1442.

\bibitem{rumelhart1988learning}
Rumelhart DE, Hinton GE, Williams RJ.
\newblock Learning representations by back-propagating errors.
\newblock Cognitive modeling. 1988;5(3):1.

\bibitem{soltoggio2008neural}
Soltoggio A.
\newblock Neural plasticity and minimal topologies for reward-based learning.
\newblock In: Hybrid Intelligent Systems, 2008. HIS'08. Eighth International
  Conference on. IEEE; 2008. p. 637--642.

\bibitem{gutstein2015reduction}
Gutstein S, Stump E.
\newblock Reduction of catastrophic forgetting with transfer learning and
  ternary output codes.
\newblock In: 2015 International Joint Conference on Neural Networks (IJCNN).
  IEEE; 2015. p. 1--8.

\bibitem{christian1990cascade}
Christian SF, Lebiere C.
\newblock The Cascade-Correlation Learning Architecture.
\newblock In: Advances in Neural Information Processing Systems 2. Citeseer;
  1990.

\bibitem{ellefsen2015neural}
Ellefsen KO, Mouret JB, Clune J, Bongard JC.
\newblock Neural Modularity Helps Organisms Evolve to Learn New Skills without
  Forgetting Old Skills.
\newblock PLoS Comput Biol. 2015;11(4):e1004128.

\bibitem{wagner2007road}
Wagner GP, Pavlicev M, Cheverud JM.
\newblock The road to modularity.
\newblock Nature Reviews Genetics. 2007;8(12):921--931.

\bibitem{guimera2005functional}
Guimera R, Amaral LAN.
\newblock Functional cartography of complex metabolic networks.
\newblock Nature. 2005;433(7028):895--900.

\bibitem{lipson2007principles}
Lipson H.
\newblock {Principles of modularity, regularity, and hierarchy for scalable
  systems}.
\newblock Journal of Biological Physics and Chemistry. 2007;7(4):125.

\bibitem{clune2013originModularity}
Clune J, Mouret JB, Lipson H.
\newblock The evolutionary origins of modularity.
\newblock Proc Royal Society B. 2013;280(20122863).

\bibitem{huizinga2014evolving}
Huizinga J, Clune J, Mouret JB.
\newblock Evolving neural networks that are both modular and regular: HyperNeat
  plus the connection cost technique.
\newblock In: Proc. Genetic \& Evolutionary Comput. Conf. ACM; 2014. p.
  697--704.


\bibitem{yao1999evolving}
Yao X.
\newblock Evolving artificial neural networks.
\newblock Proceedings of the IEEE. 1999;87(9):1423--1447.

\bibitem{levitan2015neuron}
Levitan IB, Kaczmarek LK.
\newblock The neuron: cell and molecular biology.
\newblock Oxford University Press, USA; 2015.

\bibitem{fuxe2007golgi}
Fuxe K, Dahlstr{\"o}m A, H{\"o}istad M, Marcellino D, Jansson A, Rivera A,
  et~al.
\newblock From the Golgi--Cajal mapping to the transmitter-based
  characterization of the neuronal networks leading to two modes of brain
  communication: wiring and volume transmission.
\newblock Brain research reviews. 2007;55(1):17--54.

\bibitem{agnati2006volume}
Agnati L, Leo G, Zanardi A, Genedani S, Rivera A, Fuxe K, et~al.
\newblock Volume transmission and wiring transmission from cellular to
  molecular networks: history and perspectives.
\newblock Acta Physiologica. 2006;187(1-2):329--344.

\bibitem{nicholson1998extracellular}
Nicholson C, Sykov{\'a} E.
\newblock Extracellular space structure revealed by diffusion analysis.
\newblock Trends in neurosciences. 1998;21(5):207--215.

\bibitem{bailey2000heterosynaptic}
Bailey CH, Giustetto M, Huang YY, Hawkins RD, Kandel ER.
\newblock Is heterosynaptic modulation essential for stabilizing hebbian
  plasiticity and memory.
\newblock Nature Reviews Neuroscience. 2000;1(1):11--20.

\bibitem{marder2002cellular}
Marder E, Thirumalai V.
\newblock Cellular, synaptic and network effects of neuromodulation.
\newblock Neural Networks. 2002;15(4):479--493.

\bibitem{jay2003dopamine}
Jay TM.
\newblock Dopamine: a potential substrate for synaptic plasticity and memory
  mechanisms.
\newblock Progress in neurobiology. 2003;69(6):375--390.

\bibitem{engert1997synapse}
Engert F, Bonhoeffer T.
\newblock Synapse specificity of long-term potentiation breaks down at short
  distances.
\newblock Nature. 1997;388(6639):279--284.

\bibitem{diamond2002broad}
Diamond JS.
\newblock A broad view of glutamate spillover.
\newblock Nature neuroscience. 2002;5(4):291--292.

\bibitem{schoppa2001glomerulus}
Schoppa NE, Westbrook GL.
\newblock Glomerulus-specific synchronization of mitral cells in the olfactory
  bulb.
\newblock Neuron. 2001;31(4):639--651.

\bibitem{fuxe2012extrasynaptic}
Fuxe K, Borroto-Escuela DO, Romero-Fernandez W, Diaz-Cabiale Z, Rivera A,
  Ferraro L, et~al.
\newblock Extrasynaptic neurotransmission in the modulation of brain function.
  Focus on the striatal neuronal--glial networks.
\newblock Frontiers in physiology. 2012;3.

\bibitem{baldwin2000design}
Baldwin CY, Clark KB.
\newblock Design rules: The power of modularity. vol.~1.
\newblock MIT press; 2000.

\bibitem{alon2006introduction}
Alon U.
\newblock An Introduction to Systems Biology: Design Principles of Biological
  Circuits.
\newblock CRC press; 2006.

\bibitem{carroll2001chance}
Carroll SB.
\newblock Chance and necessity: the evolution of morphological complexity and
  diversity.
\newblock Nature. 2001;409(6823):1102--1109.

\bibitem{Hintze2008}
Hintze A, Adami C.
\newblock {Evolution of complex modular biological networks.}
\newblock PLoS computational biology. 2008;4(2):e23.
\newblock doi:{10.1371/journal.pcbi.0040023}.

\bibitem{Kashtan2005}
Kashtan N, Alon U.
\newblock {Spontaneous evolution of modularity and network motifs}.
\newblock Proc Nat'l Acad Sciences. 2005;102(39):13773--13778.
\newblock doi:{10.1073/pnas.0503610102}.

\bibitem{klingenberg2005developmental}
Klingenberg CP.
\newblock {Developmental constraints, modules, and evolvability}.
\newblock Variations. 2005; p. 1--30.

\bibitem{Leicht2008}
Leicht EA, Newman MEJ.
\newblock {Community structure in directed networks}.
\newblock Physical review letters. 2008; p. 118703--118707.

\bibitem{velez2016identifying}
Velez R, Clune J.
\newblock Identifying Core Functional Networks and Functional Modules within
  Artificial Neural Networks via Subsets Regression.
\newblock In: Proceedings of the 2016 on Genetic and Evolutionary Computation
  Conference. ACM; 2016. p. 181--188.

\bibitem{hebb2005organization}
Hebb DO.
\newblock The organization of behavior: A neuropsychological theory.
\newblock Psychology Press; 2005.

\bibitem{soltoggio2007evolving}
Soltoggio A, Durr P, Mattiussi C, Floreano D.
\newblock Evolving neuromodulatory topologies for reinforcement learning-like
  problems.
\newblock In: Evolutionary Computation, 2007. CEC 2007. IEEE Congress on. IEEE;
  2007. p. 2471--2478.

\bibitem{soltoggio2008evolutionary}
Soltoggio A, Bullinaria JA, Mattiussi C, Durr P, Floreano D.
\newblock {Evolutionary Advantages of Neuromodulated Plasticity in Dynamic,
  Reward-based Scenarios}.
\newblock Artificial Life. 2008;11:569.

\bibitem{goodfellow2013empirical}
Goodfellow IJ, Mirza M, Xiao D, Courville A, Bengio Y.
\newblock An empirical investigation of catastrophic forgetting in
  gradient-based neural networks.
\newblock arXiv preprint arXiv:13126211. 2013;.

\bibitem{Kirkpatrick14032017}
Kirkpatrick J, Pascanu R, Rabinowitz N, Veness J, Desjardins G, Rusu AA, et~al.
\newblock Overcoming catastrophic forgetting in neural networks.
\newblock Proceedings of the National Academy of Sciences.
  2017;doi:{10.1073/pnas.1611835114}.

\bibitem{luders2017continual}
L{\"u}ders B, Schl{\"a}ger M, Korach A, Risi S.
\newblock Continual and One-Shot Learning Through Neural Networks with Dynamic
  External Memory.
\newblock In: European Conference on the Applications of Evolutionary
  Computation. Springer; 2017. p. 886--901.

\bibitem{ratcliff1990connectionist}
Ratcliff R.
\newblock Connectionist models of recognition memory: constraints imposed by
  learning and forgetting functions.
\newblock Psychological review. 1990;97(2):285.

\bibitem{robins1995catastrophic}
Robins A.
\newblock Catastrophic forgetting, rehearsal and pseudorehearsal.
\newblock Connection Science. 1995;7(2):123--146.

\bibitem{hinton1987using}
Hinton GE, Plaut DC.
\newblock Using fast weights to deblur old memories.
\newblock In: Proceedings of the ninth annual conference of the Cognitive
  Science Society; 1987. p. 177--186.

\bibitem{agnati2010understanding}
Agnati LF, Guidolin D, Guescini M, Genedani S, Fuxe K.
\newblock Understanding wiring and volume transmission.
\newblock Brain research reviews. 2010;64(1):137--159.

\bibitem{isaacson2000synaptic}
Isaacson JS.
\newblock Synaptic transmission: spillover in the spotlight.
\newblock Current Biology. 2000;10(13):R475--R477.

\bibitem{huang1998synaptic}
Huang EP.
\newblock Synaptic transmission: spillover at central synapses.
\newblock Current biology. 1998;8(17):R613--R615.

\bibitem{sem2005diffusional}
Sem'yanov A.
\newblock Diffusional extrasynaptic neurotransmission via glutamate and GABA.
\newblock Neuroscience and behavioral physiology. 2005;35(3):253--266.

\bibitem{kullmann1998extrasynaptic}
Kullmann DM, Asztely F.
\newblock Extrasynaptic glutamate spillover in the hippocampus: evidence and
  implications.
\newblock Trends in neurosciences. 1998;21(1):8--14.

\bibitem{scanziani2000gaba}
Scanziani M.
\newblock GABA spillover activates postsynaptic GABA B receptors to control
  rhythmic hippocampal activity.
\newblock Neuron. 2000;25(3):673--681.

\bibitem{rossi1998spillover}
Rossi DJ, Hamann M.
\newblock Spillover-mediated transmission at inhibitory synapses promoted by
  high affinity $\alpha$ 6 subunit GABA A receptors and glomerular geometry.
\newblock Neuron. 1998;20(4):783--795.

\bibitem{isaacson1999glutamate}
Isaacson JS.
\newblock Glutamate spillover mediates excitatory transmission in the rat
  olfactory bulb.
\newblock Neuron. 1999;23(2):377--384.

\bibitem{rice2011dopamine}
Rice ME, Patel JC, Cragg SJ.
\newblock Dopamine release in the basal ganglia.
\newblock Neuroscience. 2011;198:112--137.

\bibitem{descarries1996dual}
Descarries L, Watkins KC, Garcia S, Bosler O, Doucet G.
\newblock Dual character, asynaptic and synaptic, of the dopamine innervation
  in adult rat neostriatum: a quantitative autoradiographic and
  immunocytochemical analysis.
\newblock Journal of Comparative Neurology. 1996;375(2):167--186.

\bibitem{de2005synaptic}
De-Miguel FF, Trueta C.
\newblock Synaptic and extrasynaptic secretion of serotonin.
\newblock Cellular and molecular neurobiology. 2005;25(2):297--312.

\bibitem{descarries1975serotonin}
Descarries L, Alain B, Watkins KC.
\newblock Serotonin nerve terminals in adult rat neocortex.
\newblock Brain research. 1975;100(3):563--588.

\bibitem{trueta2012extrasynaptic}
Trueta C, De-Miguel FF.
\newblock Extrasynaptic exocytosis and its mechanisms: a source of molecules
  mediating volume transmission in the nervous system.
\newblock Frontiers in physiology. 2012;3:319.

\bibitem{taber2014volume}
Taber KH, Hurley RA.
\newblock Volume transmission in the brain: beyond the synapse.
\newblock The Journal of neuropsychiatry and clinical neurosciences.
  2014;26(1):iv--4.

\bibitem{wang2002two}
Wang R.
\newblock Two's company, three's a crowd: can H2S be the third endogenous
  gaseous transmitter?
\newblock The FASEB journal. 2002;16(13):1792--1798.

\bibitem{garthwaite2008concepts}
Garthwaite J.
\newblock Concepts of neural nitric oxide-mediated transmission.
\newblock European Journal of Neuroscience. 2008;27(11):2783--2802.

\bibitem{susswein2004nitric}
Susswein AJ, Katzoff A, Miller N, Hurwitz I.
\newblock Nitric oxide and memory.
\newblock The Neuroscientist. 2004;10(2):153--162.

\bibitem{esplugues2002no}
Esplugues JV.
\newblock NO as a signalling molecule in the nervous system.
\newblock British journal of pharmacology. 2002;135(5):1079--1095.

\bibitem{dawson1994gases}
Dawson TM, Snyder SH.
\newblock Gases as biological messengers: nitric oxide and carbon monoxide in
  the brain.
\newblock The Journal of neuroscience. 1994;14(9):5147--5159.

\bibitem{gally1990no}
Gally JA, Montague PR, Reeke GN, Edelman GM.
\newblock The NO hypothesis: possible effects of a short-lived, rapidly
  diffusible signal in the development and function of the nervous system.
\newblock Proceedings of the National Academy of Sciences.
  1990;87(9):3547--3551.

\bibitem{husbands1998better}
Husbands P, Smith T, Jakobi N, O'Shea M.
\newblock Better living through chemistry: Evolving GasNets for robot control.
\newblock Connection Science. 1998;10(3-4):185--210.

\bibitem{mchale2004gasnets}
McHale G, Husbands P.
\newblock Gasnets and other evolvable neural networks applied to bipedal
  locomotion.
\newblock From Animals to Animats. 2004;8:163--172.

\bibitem{rice2008dopamine}
Rice ME, Cragg SJ.
\newblock Dopamine spillover after quantal release: rethinking dopamine
  transmission in the nigrostriatal pathway.
\newblock Brain research reviews. 2008;58(2):303--313.

\bibitem{rusakov1998extrasynaptic}
Rusakov DA, Kullmann DM.
\newblock Extrasynaptic glutamate diffusion in the hippocampus: ultrastructural
  constraints, uptake, and receptor activation.
\newblock The Journal of neuroscience. 1998;18(9):3158--3170.

\bibitem{wood1994models}
Wood J, Garthwaite J.
\newblock Models of the diffusional spread of nitric oxide: implications for
  neural nitric oxide signalling and its pharmacological properties.
\newblock Neuropharmacology. 1994;33(11):1235--1244.

\bibitem{mengistu2016evolutionary}
Mengistu H, Huizinga J, Mouret JB, Clune J.
\newblock The evolutionary origins of hierarchy.
\newblock PLoS Comput Biol. 2016;12(6):e1004829.

\bibitem{fine2006spatially}
Fine P, Di~Paolo E, Philippides A.
\newblock Spatially constrained networks and the evolution of modular control
  systems.
\newblock In: International Conference on Simulation of Adaptive Behavior.
  Springer; 2006. p. 546--557.

\bibitem{witten1981diffusion}
Witten~Jr T, Sander LM.
\newblock Diffusion-limited aggregation, a kinetic critical phenomenon.
\newblock Physical review letters. 1981;47(19):1400.

\bibitem{stanley2007compositional}
Stanley KO.
\newblock {Compositional pattern producing networks: A novel abstraction of
  development}.
\newblock Genetic Programming and Evolvable Machines. 2007;8(2):131--162.

\bibitem{risi2010indirectly}
Risi S, Stanley K.
\newblock {Indirectly Encoding Neural Plasticity as a Pattern of Local Rules}.
\newblock From Animals to Animats 11. 2010; p. 533--543.

\bibitem{rusu2016progressive}
Rusu AA, Rabinowitz NC, Desjardins G, Soyer H, Kirkpatrick J, Kavukcuoglu K,
  et~al.
\newblock Progressive neural networks.
\newblock arXiv preprint arXiv:160604671. 2016;.

\bibitem{mouret2010sferes}
Mouret JB, Doncieux S.
\newblock Sferes v2: Evolvin'in the multi-core world.
\newblock In: Evolutionary Computation (CEC), 2010 IEEE Congress on. IEEE;
  2010. p. 1--8.

\bibitem{tonelli2011relationships}
Tonelli P, Mouret JB.
\newblock On the relationships between synaptic plasticity and generative
  systems.
\newblock In: Proceedings of the Genetic and Evolutionary Computation
  Conference. ACM; 2011. p. 1531--1538.

\bibitem{deb2002fast}
Deb K, Pratap A, Agarwal S, Meyarivan T.
\newblock A fast and elitist multiobjective genetic algorithm: NSGA-II.
\newblock IEEE transactions on evolutionary computation. 2002;6(2):182--197.

\bibitem{Mouret2012}
Mouret JB, Doncieux S.
\newblock Encouraging Behavioral Diversity in Evolutionary Robotics: an
  Empirical Study.
\newblock Evolutionary Computation. 2012;1(20).

\bibitem{2001multi}
Deb K.
\newblock Multi-objective optimization using evolutionary algorithms. vol.~16.
\newblock Wiley; 2001.

\end{thebibliography}
\end{document}